\pdfoutput=1
\documentclass[10pt]{article}

\usepackage[T1]{fontenc}
\usepackage[margin=1in]{geometry}
\usepackage[hidelinks,pdfauthor={Lngel Kyo},pdftitle={},pdfsubject={},pdfkeywords={}]{hyperref}
\usepackage{url}
\usepackage{booktabs}
\usepackage{array}
\usepackage{amsfonts}
\usepackage{microtype}
\usepackage{amsmath, amssymb, amsthm}
\usepackage{graphicx}
\usepackage{natbib}
\usepackage{flafter}
\makeatletter
\g@addto@macro\@listi{\itemsep=2pt\parsep=0pt\topsep=3pt}
\makeatother
\newcommand{\R}{\mathbb{R}}
\newcommand{\Sig}{\boldsymbol{\Sigma}}
\newcommand{\sstar}{s^{\ast}}
\newtheorem{proposition}{Proposition}

\title{Breaking Precision Walls in First-, Second-, and Zeroth-Order\\
Optimizers: One Constant, Set in Advance}

\author{Lngel Kyo\\
  A2O Labs\\
  \texttt{lngel.kyo@a2o-labs.com}}
\date{}

\begin{document}
\vspace{-2.2em}
\maketitle
\vspace{-2.2em}

\begin{abstract}
When signal strength decays across directions as a power law, an optimizer can fit strong directions and silently miss weak ones while an average loss hides it. We measure this with a \emph{precision wall}: the steepest decay at which an optimizer still recovers the weakest direction to within 5\% in a fixed budget. On low-rank matrix sensing and quadratics one constant sets the wall: once it dominates the denominator, Adam or grafted Shampoo is momentum gradient descent with step $\eta / c$, and the wall follows this ratio; in Shampoo, matching the grafting constant to the damping restores the wall after three earlier hypotheses failed. In zeroth-order Adam, $\varepsilon$ acts against $\varepsilon^\ast$ computed from the spectrum by a mean-square model; below $\varepsilon^\ast$ the error settles on a noise floor. Set by this rule from the true curvature or spectrum on new shapes, the constant raised the wall above the same-step default by a forecast amount: on two matrix-sensing shapes by $1.96$ and $1.15$ for Adam and $1.59$ and $1.74$ for grafted Shampoo, and on two Hadamard-rotated quadratics by $1.12$ and $0.73$ for zeroth-order Adam at the longer budget, rising on every seed. Fixed-step momentum methods with the same step reach the same walls. A cosine-scheduled default, whose small constant gives weak directions large steps, goes higher on one matrix-sensing shape, is matched on new seeds of the other, and falls when combined with the rule's constant. A default step size retuned per budget catches up on the rotated quadratics and population logistic regression, does not at one query per step on two diagonal quadratics and does at sixteen, and exceeds the rule on handwritten-digits logistic regression (full-batch and under minibatches, where gradient noise sets the $\eta = 0.01$ and $0.1$ default walls). On logistic regression the rule's constant did not raise the first-order wall: same-step default Adam was higher, its wall on four new shapes falling on the forecast stability limit, whereas in zeroth order on the population loss changing only $\varepsilon$ raised the wall by $0.98$ and $0.88$.
\end{abstract}

\section{Introduction}
\label{sec:intro}

Optimizer benchmarks usually report an average: validation loss, or, as in AlgoPerf~\citep{dahl2023benchmarking,kasimbeg2025accelerating}, the time needed to reach a validation target. When the data follow a power law, as word frequencies do~\citep{kunstner2024zipf} and as many feature spectra do~\citep{bordelon2024dynamical}, that average is set by a few strong directions. It says little about whether the rare directions were learned at all.

To measure that gap directly, we take a family of problems in which the signal decays across directions, turn up the decay, and record the steepest decay at which an optimizer still recovers the weakest direction to within 5\% in a fixed budget: its precision wall. We expect a wall wherever a weak signal moves the parameters too slowly for the budget or cannot get past a floor. In an adaptive method the small constant added to a denominator can slow it. In a method that estimates gradients from loss values alone, the noise of the estimate can set the floor.

Our main result is that the wall can be raised by a forecast amount (Section~\ref{sec:break}), because one constant sets it. Once the small constant $c$ in a denominator dominates, an update of Adam~\citep{kingma2015adam} or of grafted Shampoo is momentum gradient descent with step $\eta / c$, and the weakest direction closes at a rate set by that step ratio (Section~\ref{sec:step}). In zeroth-order Adam, which estimates gradients from pairs of loss values, one random sign vector per step collapses Adam's per-coordinate denominator to a single number; below a value $\varepsilon^\ast$ of its constant the error settles on a noise floor, and a mean-square model computes $\varepsilon^\ast$ from the spectrum (Section~\ref{sec:zo}). Setting the constant by these rules before any run, on problems not run before, raised the wall above the default with the same step size on every seed, by amounts forecast in advance: for Adam and grafted Shampoo on two shapes of matrix sensing, and for zeroth-order Adam on two quadratics turned by a Hadamard rotation with random signs. Fixed-step momentum methods with the same step reach the same walls, so what raises the wall is the step that the constant sets, not the normalization.

Where the rule raises the wall and where it does not follow from the same model, which also forecasts the default walls there. The first-order constant can come from the measurements alone on this problem at a precomputed cost; in zeroth order, a starting-trace constant without the spectrum holds under the conditions in Section~\ref{sec:further}. A cosine-scheduled default, whose small constant gives weak directions large steps, goes higher than the rule on one matrix sensing shape and is matched on new seeds of the other; adding the rule's constant lowers its wall. On the rotated quadratics and population logistic regression a default step size retuned per budget caught up, on two diagonal quadratics it did not at one query per step; the rule needs no retuning. Over longer budgets the zeroth-order walls follow the model, capped under the default $\varepsilon$ only. On matrix sensing, whose loss is not quadratic, the rule's $\varepsilon$ removes the noise floor and stops on a second floor from the cubic term of the two-point difference. On population logistic regression the first-order default already sits at the forecast stability limit, so the first-order rule did not raise the wall while the zeroth-order one did; on handwritten digits it met its forecast full-batch without exceeding the default (nor under minibatches, where gradient noise sets the $\eta = 0.01$ and $0.1$ default walls as forecast in Table~\ref{tab:minibatch}, within $0.25$ of the highest default, and scaling $\eta$ and $c$ above the noise did not restore the quadratic wall), and a larger default step was higher. The evidence is matrix sensing (two shapes in first order, one in zeroth order), diagonal or Hadamard-rotated quadratics, and population and digits logistic regression.

The rule grew out of a study of Shampoo~\citep{gupta2018shampoo,anil2020scalable,shi2023distributed}, which adds damping to its preconditioners. Raising the damping lowers Shampoo's wall, and grafting~\citep{agarwal2020disentangling} with Adam's constant matched to the square root of the damping puts the median wall back on the ungrafted one (Section~\ref{sec:main}). Three earlier studies failed (Section~\ref{sec:failed}): moving from Shampoo to SOAP lowered the wall, cutting a preconditioner to its diagonal did not collapse it, and a larger population did not raise the wall of Evolution Strategies~\citep{salimans2017evolution}.

Section~\ref{sec:setup} defines the problem and wall. Sections~\ref{sec:failed}--\ref{sec:zo} report the failed hypotheses, Shampoo study, first-order rule and zeroth-order model, tested on runs fixed in advance; Sections~\ref{sec:break}--\ref{sec:beyond} raise the wall on new problems and map where it does not rise, before discussion, related work and limitations.

\section{Problem and Measurement}
\label{sec:setup}

\textbf{The problem.}
We recover a low-rank positive semidefinite matrix from linear measurements. Given $V$ and rank $D \le V$, we observe $m = \mathrm{round}(1.5\,VD)$ measurements $b_k = \langle A_k, M^\ast \rangle$ of a rank-$D$ target $M^\ast$, where $A_k = (G_k + G_k^\top) / 2$ and $G_k$ has independent $\mathcal{N}(0, 1/(Vm))$ entries. We train a factor $U \in \R^{V \times D}$~\citep{burer2005local} by minimizing
\begin{equation}
\label{eq:loss}
\mathcal{L}(U) = \frac{1}{m}\sum_{k=1}^m \bigl(\langle A_k, U U^\top \rangle - b_k\bigr)^2
\end{equation}
from $U_0 = 0.1 \cdot \mathrm{randn}(V, D)$, with $T = 60{,}000$ full-batch double-precision steps. The target's eigenvalues follow a power law rotated by a random orthogonal $Q$:
\begin{equation}
\label{eq:target}
M^\ast = Q\,\mathrm{diag}\bigl(1^{-s}, 2^{-s}, \dots, D^{-s}, 0, \dots, 0\bigr)\,Q^\top.
\end{equation}
At $s = 0$ all $D$ directions have equal strength; larger $s$ weakens the last direction.

\begin{figure}[t]
\centering
\includegraphics[width=0.64\textwidth]{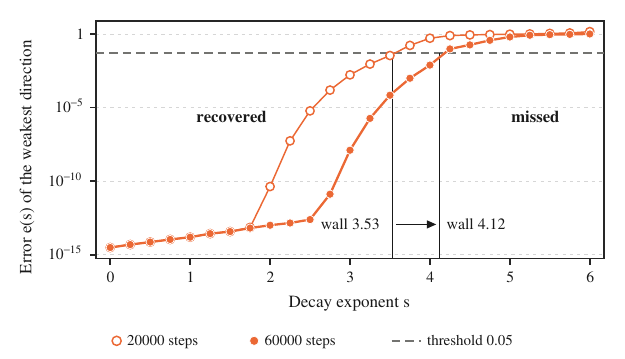}
\caption{Reading a wall: ungrafted Shampoo at damping $10^{-7}$ on $(V, D) = (12, 4)$, median of five seeds. A larger budget moves the crossing of $0.05$ from $3.53$ to $4.12$.}
\label{fig:wall}
\end{figure}

\textbf{The wall.}
We track the weakest direction $q_D$ of the target through its relative residual after $T$ steps, $e(s) = \|(U_T U_T^\top - M^\ast)\,q_D\|_2 / D^{-s}$. At each $s$, five seeds give the median or mean of $e(s)$. The \emph{precision wall} $\sstar$ is the first point where that curve crosses $0.05$, interpolated linearly between adjacent points of a grid $s_0 < \dots < s_K$ and, where noted, sharpened by bisection (Figure~\ref{fig:wall}). When a curve crosses $0.05$ once, first and last crossings coincide; when it recrosses, the rule reads the first crossing and defaults in Sections~\ref{sec:break}--\ref{sec:beyond} read the last (every minibatch setting reads the last), a pairing that never favors the rule since the first crossing never exceeds the last; when $e(s_0) > 0.05$ at the left edge (as on six curves in Section~\ref{sec:beyond}), both first upward and last crossings equal $s_0$. A diverging run counts as error $1$; a wall at the right grid edge ($s = 6$ in Sections~\ref{sec:failed}--\ref{sec:main}, $4$ in Section~\ref{sec:zo}) means no crossing was found.

\textbf{How each study was run.}
Each registered test fixed its procedure, step sizes and bands on a small calibration problem before running a larger test problem once (after reading the first test in Section~\ref{sec:main}; Section~\ref{sec:step} committed forecasts beforehand; later checks are noted).

\textbf{Optimizers.}
The reference is Adam~\citep{kingma2015adam}, with step size $\eta = 0.01$ and moment decays $\beta_1 = 0.9$ and $\beta_2 = 0.999$, shared by all methods unless stated. Adam divides by $\sqrt{\hat{v}} + \varepsilon$, so $\varepsilon$ takes over when gradients fall to about $\varepsilon$~\citep{everett2024scaling}. Shampoo keeps running averages $L$ and $R$ of $GG^\top$ and $G^\top G$ for the gradient $G$, bias-corrected like $\hat{v}$, recomputes its inverse roots every 10 steps, and moves along $(L + \varepsilon I)^{-1/4}\, \hat{m}\, (R + \varepsilon I)^{-1/4}$, where $\hat{m}$ is the bias-corrected momentum. Damping $\varepsilon$ is added to squared gradients, so below $\sqrt{\varepsilon}$ the update is the momentum divided by $\sqrt{\varepsilon}$, Shampoo's effective denominator~\citep{eschenhagen2026clarifying}. Multiplying the loss by $c$ is the same as dividing Shampoo's damping by $c^2$ and Adam's constant by $c$. SOAP~\citep{vyas2025soap} runs Adam in the eigenbasis of $L$ and $R$, and Eigen-Adafactor runs Adafactor~\citep{shazeer2018adafactor} in that basis. Grafting rescales each Shampoo update so that its Frobenius norm equals that of an Adam update computed alongside, with Adam's constant at $10^{-8}$ unless stated.

\section{Three Hypotheses That Failed}
\label{sec:failed}

Three studies came first, and each failed on its test problem. Appendix~\ref{app:failed} gives their walls in one table.

\textbf{Does the wall rise on the way from Shampoo to SOAP?}
We expected the wall to rise at both steps of the path from Shampoo through Eigen-Adafactor to SOAP, at one shared damping of $10^{-8}$. It fell by $0.7605$ in total on the problem $(V, D) = (16, 4)$. At that setting Shampoo's effective denominator is $10^{-4}$ against a constant of $10^{-8}$ in the others, so one shared number is not one shared denominator, and our Shampoo study sweeps the damping for that reason.

\textbf{Does cutting SOAP's left preconditioner to its diagonal collapse the wall?}
We expected a large drop as the matrix was truncated to smaller and smaller diagonal blocks. The wall barely moved: the full matrix's wall sat $0.1789$ below the diagonal's, read on a coarse grid. On $(16, 4)$, diagonalizing this matrix does not create a wall.

\textbf{Does the wall of Evolution Strategies grow with its population at the predicted rate?}
We expected a heuristic formula to predict the gain in the wall as the population grew from 4 to 128 pairs. Under the step-size schedule we had fixed in advance, the measured gain was $-0.1676$ times the predicted one. That schedule also raised the step size, and in an unscored check with the step size held fixed the gain returned to the predicted range, so zeroth-order scaling studies have to vary step size and sample count separately.

\section{Damping and Grafting in Shampoo}
\label{sec:main}

The first failed study showed that Shampoo's damping acts on a different scale from Adam's, so we swept it over nine levels, $\log_{10}\varepsilon = -8.0, -7.5, \dots, -4.0$, in two tests. The first ran one problem, $(V, D, m) = (12, 4, 72)$; its main statistic did not reproduce. The second uses new seeds on $(12, 4)$ and a new shape, $(16, 6, 144)$, which carries the test's weight. It has four arms on one grid of $s$ with step $0.25$, without bisection: Shampoo; grafted Shampoo with Adam's constant at $10^{-8}$ (the \emph{fixed} arm) or at $\sqrt{\varepsilon}$ (the \emph{matched} arm); and the reference, Adam at constant $10^{-8}$. Pairs below give $(12, 4)$, then $(16, 6)$, from the second test unless stated.

\begin{figure}[t]
\centering
\includegraphics[width=\textwidth]{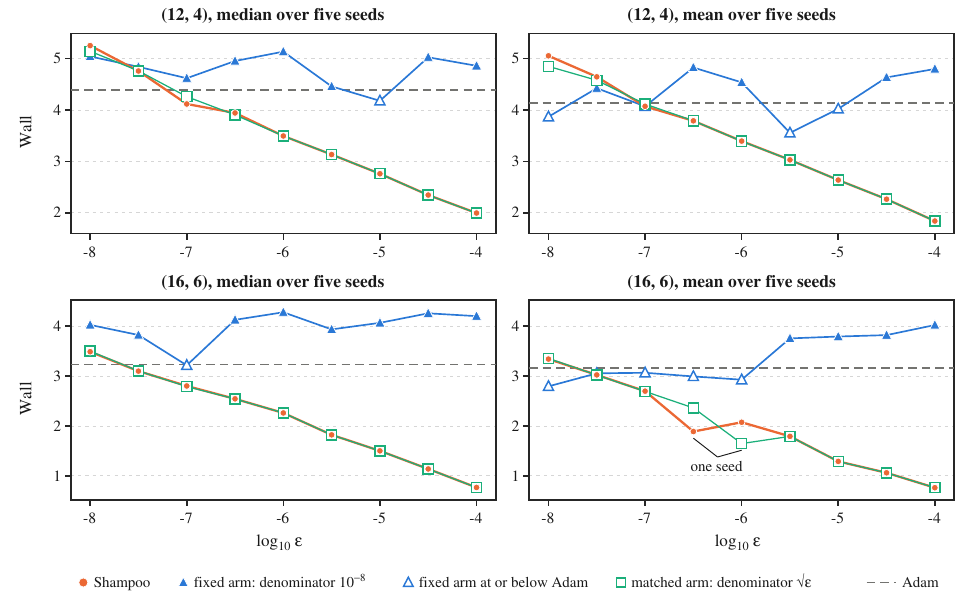}
\caption{Walls in the second test. Open triangles mark the fixed arm at or below Adam's wall, on the median at $-5.0$ and $-7.0$. The gap between circles and squares at $-6.5$ and $-6.0$, lower right, comes from isolated excursions of one seed.}
\label{fig:matched}
\end{figure}

\textbf{The registered test.}
When the constants dominate both denominators, Shampoo's step is the momentum over $\sqrt{\varepsilon}$ and Adam's the momentum over its constant, so with that constant at $\sqrt{\varepsilon}$ the matched update equals the ungrafted one by construction. The test asks whether that regime decides the wall. The rule, registered before any run but after reading the first test, reads the median over five seeds: averaged over the nine levels, the matched wall must lie within $0.0625$ of the ungrafted wall and the fixed wall at least $0.75$ above it. Both problems met both bands (Figure~\ref{fig:matched}). On average the matched median wall lies $0.033$ and $0.003$ from the ungrafted one, and the matched mean wall $0.036$ and $0.10$, where $0.10$ is above the band and comes from two isolated excursions of one seed; without that seed, checked afterwards, it is $0.0025$. Adding $15$ seeds cuts the mean-wall gaps on $(16, 6)$ at $-6.5$ and $-6.0$ from $0.4732$ and $0.4271$ to $0.0101$ and $0.0003$; the registered readings stand. The fixed arm sits on average $1.26$ and $1.83$ above on the median and $0.89$ and $1.37$ on the mean. Read afterwards, the registered test's median still meets both bands at $20000$ steps, at thresholds $0.02$ and $0.1$, and with any one seed left out. The mean does not: the matched gap of $0.10$ appears only at $60000$ steps and thresholds up to $0.05$, and the fixed arm's mean gap on $(12, 4)$ falls to $0.20$ at threshold $0.02$.

The median gap leans on the three lowest levels, since the other six are near zero by construction: on $(12, 4)$ those three average $0.0858$, above the band, and the largest gap is $0.14$, against $0.011$ on $(16, 6)$. With $20$ seeds the average falls to $0.0325$: five seeds can explain its size, but it does not reach zero, and the matched arm is higher at all three levels. What agrees is the wall and not the run: near the wall, errors at the same $s$ and seed agree to $1\%$ at the highest levels but at few points at the lowest.

\textbf{What we also observed.}
Three patterns, seen on both problems, lay outside any rule. The ungrafted median wall falls at every level, from $5.25$ to $1.99$ and from $3.49$ to $0.77$, and the mean wall falls too, but for one rise at that seed's excursions. At this budget it first drops below this Adam reference at $-7.0$ and $-7.5$, on median and mean. The fixed arm does not follow the damping down: from $-8.0$ to $-4.0$ the ungrafted median wall drops by $3.26$ and $2.72$, while the fixed arm's changes by $-0.18$ and $+0.18$ and its mean wall rises by $0.93$ and $1.23$. It sits above the ungrafted arm at $17$ of $18$ levels on the median and $14$ on the mean, every exception at low damping.

\textbf{What did not hold.}
The first test set out to measure how far grafting moves the crossing where the wall first falls below Adam's. There the grafted median wall stayed above Adam's at all nine levels, so the shift read $+3.0$, set by the grid end, and the mean read $+1.0$. In the second test it reads $+2.0$ and $-1.0$ on $(12, 4)$ and $+0.5$ and $-0.5$ on $(16, 6)$, median then mean; the first test's rule counted $+0.5$ or less as failed. The fixed arm scatters: $38$ and $29$ of its $45$ seed curves reach the threshold and come back, against $0$ and $1$ without grafting (under last crossing, $9$ of $18$ fixed-arm median walls move, up to $1.4749$; this section keeps first crossing). Scatter sets where it first touches Adam's: on $(16, 6)$ by $0.0155$, with the next level $0.90$ above. Its median wall exceeds Adam's at $16$ of $18$ levels and its mean at $9$. We therefore report no shift.

\textbf{How far to trust it.}
Since $(12, 4)$ repeats a problem we had run, only $(16, 6)$ could have overturned the test. The test compared only two denominator values. Falling below Adam's holds for one reference: of the first test's eleven Adam configurations, six gave the same shape and five fell in the failed range. Refreshing the preconditioner every step left the ungrafted crossing unchanged, measured only in the first test. The two grids differ, so we do not compare walls; Appendix~\ref{app:first} shows the first.

\section{One Constant Sets the Wall}
\label{sec:step}

Once a constant $c$ dominates the denominator, an update is gradient descent with momentum and step $\eta / c$: $c = \sqrt{\varepsilon}$ for ungrafted Shampoo, Adam's constant for Adam, and the grafting constant $\varepsilon_g$ for grafted Shampoo when the damping dominates. The weakest direction then closes at the rate $(\eta / c)\, h_{\min}$, where $h_{\min}$ is the instance's smallest curvature at the solution outside the rotation gauge. An error starting at $1$ reaches $0.05$ after $\ln 20$ e-folds, so on each seed the wall is the $s$ where
\begin{equation}
\label{eq:step}
\frac{\eta}{c}\, h_{\min}(s)\, T = \ln 20 ,
\end{equation}
and the forecast is the median over seeds (Section~\ref{sec:discussion} evaluates adding the non-zero initial log-error $\ln e_0$). The rule needs $A = (\eta / c)\, h_{\max}$, the step times the largest curvature, below $38$, the stability limit of momentum at $\beta_1 = 0.9$~\citep{cohen2022adaptive}: under a small default constant, the fastest mode crosses $L = 2(1 + \beta_1)/(1 - \beta_1) = 38$ into period-two oscillation and, through measurement coupling, pins the weakest direction's denominator at a self-generated floor much larger than $10^{-8}$~\citep{pw2026a,pw2026b}; raising $c$ to $\eta\, h_{\max}/36$ suppresses that oscillation so weak directions relax at rate $(\eta / c)\,h_{\min}$.

\textbf{Status.}
These supplementary readings stand alone. After reading earlier runs, we tested the rule on new seeds and a shape not yet run, $(14, 5)$, committing forecasts, reader and bands before any run with no fitted constant (bands set after the earlier runs); a separate agent ran the plans and a program read each outcome once. Walls are five-seed medians on a $0.25$ grid, interpolated in $\ln(-\ln e)$; one is scored if its forecast lies in $[1.5, 5]$ with $A < 38$ on every seed.

\begin{figure}[t]
\centering
\includegraphics[width=\textwidth]{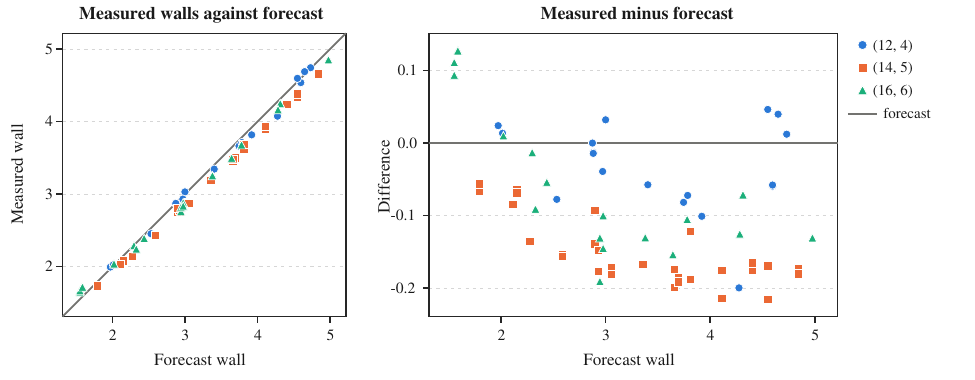}
\caption{The $68$ scored walls against their forecasts. The differences, right, lean below zero.}
\label{fig:step}
\end{figure}

\begin{table}[ht]
\centering
\small
\caption{Measured minus forecast on the scored walls. Bands: rms at most $0.20$, largest at most $0.50$; $0.25$ and $0.60$ for the ladder and step sizes.}
\label{tab:step}
\begin{tabular}{lrrrr}
\toprule
Walls & Number & Below & rms & Largest \\
\midrule
Ungrafted Shampoo, $(14, 5)$ & $16$ & $16$ & $0.1637$ & $0.2155$ \\
Adam at $\sqrt{\varepsilon}$, $(14, 5)$ & $16$ & $16$ & $0.1481$ & $0.1915$ \\
Grafted, ladder of $\varepsilon_g$ & $10$ & $8$ & $0.0917$ & $0.1547$ \\
Ungrafted, other step sizes & $5$ & $3$ & $0.1348$ & $0.1918$ \\
Ungrafted, long budgets & $13$ & $10$ & $0.0876$ & $0.1460$ \\
Ungrafted, measurement ratios & $8$ & $3$ & $0.0809$ & $0.1998$ \\
\midrule
All & $68$ & $56$ & $0.1276$ & $0.2155$ \\
\bottomrule
\end{tabular}
\end{table}

\textbf{Walls land below the forecast.}
Over three shapes $68$ walls were scored (Figure~\ref{fig:step}, Table~\ref{tab:step}; instances recur across studies, so these are not independent). The rms difference is $0.1276$ and the largest $0.2155$, and $56$ of the $68$ lie below the forecast, including $45$ of the $50$ forecast above $2.5$. On $(14, 5)$ all $32$ lie below, by $0.0562$ to $0.2155$, so there the rms is nearly all bias; on $(12, 4)$ and $(16, 6)$, $11$ and $13$ of $18$ do, and the new shape shows the largest bias. All six lines met their size bands; Section~\ref{sec:discussion} compares the late-checkpoint decay rate with $(\eta / c)\, h_{\min}$ and reads the early-step offset.

\textbf{Budget.}
The wall rises with the budget, by close to the forecast: on $(14, 5)$, from $60000$ to $240000$ steps, by $0.8653$ on average against $0.9164$. On the same instances, four other step sizes with the same $\eta T$ move the wall by at most $0.0060$, and at step size $0.01$ Adam at $\sqrt{\varepsilon}$ differs from Shampoo at $\varepsilon$ by rms $0.0273$.

\textbf{Grafting.}
When the damping dominates, the grafted wall follows $\varepsilon_g$: ten walls on a ladder of $\varepsilon_g$ differ from the forecast with $c = \varepsilon_g$ by rms $0.0917$, eight below it, and moving the damping from $10^{-4}$ to $10^{-6}$ moves three walls by at most $0.0410$. At step sizes $0.003$ and $0.03$ on $(16, 6)$ the matched arm stays close to the ungrafted arm, with mean gaps $0.0070$ and $0.0371$. The fixed arm's curves are flat and ragged near the threshold, so we quote only signs: its wall and that of Adam at $10^{-8}$ fall as the step rises, and at damping $10^{-8}$ it sits above the ungrafted arm at $\eta = 0.003$ and below at $\eta = 0.03$. From $60000$ to $240000$ steps the ungrafted wall rises $0.6696$ to $0.9801$, the fixed arm does not rise with it, and their gap shrinks in all four settings, so grafting's lift depends on the budget, not a fixed difference between methods. One reading stayed undecided: an intermediate $\varepsilon_g$ above both ends was read only on $(16, 6)$; on $(12, 4)$ the level named in advance did not rise above the fixed arm, and we claim nothing from that half.

\textbf{The limit.}
The scored walls reach $A$ at most $4.39$ and do not test $38$. A study on $(14, 5)$ set $c$ on each instance to reach a chosen $A$, for Adam and for Shampoo refreshed every step. At $A \le 36$ the error fell below $10^{-9}$ in all $45$ rows of each arm; at $A = 50$ and $60$ it stayed above in $25$ of each arm's $30$. The boundary lies between the values we ran, $36$ and $40$. The arms agree on $84$ of $105$ single rows, so $A$ does not decide one run, and SOAP was not in this study.

\textbf{Measurement ratio.}
On $(12, 4)$ at $60000$ steps, with $m / (VD)$ from $1.25$ to $4$, eight walls differ from the instance's forecast by rms $0.0809$, three below it. All eight are closer than the one alternative named in advance, which keeps only the generator's $1 / m$ scale. The wall rises and falls with $m$, like the forecast.

\section{Zeroth-Order Adam: A Mean-Square Model of the Wall, Tested in Advance}
\label{sec:zo}

Zeroth-order methods replace the gradient with an estimate built from loss values, as MeZO does~\citep{malladi2023fine}. With one random sign vector $z \in \{\pm 1\}^{VD}$ per step, the two-point estimate is $\hat{g} = a\,z$ for a single scalar $a$, so every coordinate has the same square $a^2$. Feeding this estimate to Adam makes the second-moment estimate $\hat{v}_t$ one number shared by all coordinates, and the method reduces to momentum SGD divided by a single scalar. Its step on Adam's first-moment estimate is $\gamma_t = \tilde{\eta} / (\sqrt{\hat{v}_t} + \varepsilon)$, where $\tilde{\eta}$ is the step-size setting. This is an algebraic identity, since $z_i^2 = 1$, and not a finding. For Gaussian and spherical perturbations, \citet{dbouk2026adaptivity} approximate the expected coordinate-wise square by a gradient-norm term plus a per-coordinate term, and the first dominates when $q \ll d$ perturbations are drawn per step and $\|\nabla f\|^2$ grows with $d$ faster than $1 / q$. The exact form for sign vectors is our own derivation. The constant $\varepsilon$ sets the scale below which the division stops and the method becomes plain momentum SGD with step $\tilde{\eta} / \varepsilon$.

\textbf{Where the step settles.}
A scalar normalizer is known to push the effective step to the largest stable step. \citet{khaled2023dowg} observe this in a deterministic experiment with normalized gradient descent on regularized linear regression: the effective step oscillates around $2 / L$, with $L$ the smoothness constant, and they argue that a step past that threshold makes the gradients grow, which forces the step back down. \citet{cohen2025understanding} show the same for deterministic full-batch scalar RMSProp, where along the time-averaged trajectory the effective step equals $2 / S(w)$, with $S(w)$ the sharpness, and \citet{regis2026rod} show the preconditioned sharpness of scalar RMSProp hovering around its threshold $2 / \eta$. For zeroth-order Adam, \citet{song2026zeroth} report the experimental relation $\mathrm{Tr}(P^{-1} H) \approx 2 / \eta$ at the mean-square edge of stability, with $P$ the preconditioner, and their analysis freezes $P$. We give the mean-square counterpart for sign-vector estimates on a quadratic. Let the loss be $\tfrac{1}{2} \sum_i \lambda_i e_i^2$, with $e = \theta - \theta^\ast$ and Hessian $H = \mathrm{diag}(\lambda_i)$, and write $A_i = \mathbb{E}[e_i^2]$ and $G^2 = \mathbb{E}[a^2] = \sum_i \lambda_i^2 A_i$, the quantity that $\hat{v}_t$ averages.

\begin{proposition}
\label{prop:step}
Suppose the loss is a pure quadratic with Hessian diagonal in the coordinates of $z$, the two-point difference carries no minibatch noise, and each step uses one sign vector. Then for any fixed sequence of steps $\gamma_t$ the second moments of the error and of the first moment obey an exact closed recursion. At a constant step this recursion has a stationary state with $G > 0$ only at $\gamma = \gamma^\ast$, the root of
\begin{equation}
\label{eq:gstar}
\frac{\gamma}{2} \sum_i \frac{\lambda_i}{1 + \kappa \gamma \lambda_i} = 1,
\qquad \kappa = \frac{\beta_1}{1 + \beta_1},
\end{equation}
and in that state
\begin{equation}
\label{eq:floor}
A_i = \frac{\gamma^\ast G^2}{2 \lambda_i \, (1 + \kappa \gamma^\ast \lambda_i)}.
\end{equation}
At $\beta_1 = 0$, $\gamma^\ast = 2 / \mathrm{tr}\,H$; for $\beta_1 \in [0, 1)$, the second moments contract to zero below $\gamma^\ast$ and diverge above it.
\end{proposition}

Appendix~\ref{app:prop} derives Proposition~\ref{prop:step}: at a constant step the four-line recursion is a $3d$-dimensional linear map on $(A_i, B_i, C_i)$ coupled through $G^2$, and mean-square stability is its spectral radius lying below $1$. At $\beta_1 = 0$, a two-point step on a quadratic is noiseless least mean squares after a change of variables, whose largest stable step \citet{defossez2015averaged} bound by $2 / \mathrm{tr}\,H$ with equality for constant-norm sign vectors (Appendix~\ref{app:prop}); for Gaussian directions, $2 / \mathrm{tr}\,H$ is the upper bound in Theorem~1 of \citet{song2026zeroth}.

Going back to Adam's own step takes one approximation, so what follows is a mean-field statement and not part of the proposition: we replace $\hat{v}_t$ by an average of $G^2$ and ignore its fluctuations, needing a $\beta_2$ window longer than the strongest modes' mixing time. If $\gamma_t < \gamma^\ast$ the error shrinks, $\hat{v}_t$ falls and $\gamma_t$ rises; if $\gamma_t > \gamma^\ast$ the reverse happens. So with $\varepsilon$ negligible the method raises its own step to the mean-square stability limit $\gamma^\ast$, where $\sqrt{\hat{v}_t} = \tilde{\eta} / \gamma^\ast$, and the error stops at the level of \eqref{eq:floor} instead of falling to zero. We call this level the noise floor. The formula applies only after the gradient scale, and $\hat{v}_t$ with it, has come down from its large early value to the stationary level, and after the budget exceeds the slowest relaxation time $1 / (2 \gamma^\ast \lambda_{\min})$; with a shorter budget only the recursion applies. With $\varepsilon$ put back, the stationary state needs $\tilde{\eta} / (G + \varepsilon) = \gamma^\ast$, so $G = \varepsilon^\ast - \varepsilon$ with $\varepsilon^\ast := \tilde{\eta} / \gamma^\ast$. Below $\varepsilon^\ast$ the error at the floor shrinks by the factor $1 - \varepsilon / \varepsilon^\ast$; from $\varepsilon^\ast$ on there is no stationary floor, and only convergence is left. By the model, $\varepsilon^\ast$ can be read without knowing $H$, as the stationary $\sqrt{\hat{v}_t}$ of a run with $\varepsilon \to 0$. A wall is thus set either by the floor, independent of budget, or by convergence. Moment equations at finite step size have also been written for minibatch SGD. \citet{schmid2026stochastic} close them at second order with a Gaussian closure, and in the full-batch limit the edge of stability appears as a stationary point of the sharpness and covariance dynamics. Their general equations cover any stochastic update rule $\theta \to \theta + \eta f(\theta)$, but they work out only SGD, and neither a normalizer nor direction sampling appears in their analysis.

\textbf{Test problems.}
Both tests use a quadratic on $\Theta \in \R^{V \times D}$ with curvature $\lambda_{rc} = \bigl(1 + 0.22\,(r-1) + 0.08\,(c-1)\bigr)^{-s}$, evaluated without noise. Each run starts from $\theta_0 = \theta^\ast + 0.75 \cdot \mathrm{randn}(V, D)$, where $\theta^\ast$ has entries of random sign and magnitude $1 + 0.12 \cdot \mathrm{randn}$. The tests measure the median over five seeds of the error on the last three quarters of the rows, relative to its value at the start, after $T = 156.25\,VD$ steps, on a grid of $s$ from $1$ to $4$ with step $0.25$ and four bisections, taking the lower end of the bisection bracket at the first crossing. Settings come from the calibration problem $(32, 4)$ and run on $(64, 4)$, $(64, 8)$, $(128, 4)$ and $(128, 8)$, with $\tilde{\eta}$ shrinking in proportion to $1 / (1 + VD)$.

\textbf{A first test.}
It asked whether raising $\varepsilon$ moves the wall once each setting has tuned its own step size. One setting kept the default $\varepsilon = 10^{-8}$; the other selected $\varepsilon = 0.01$ and about twice the default step. A calibration check stopped it once, and we recorded a changed check before rerunning. Only threshold $0.05$ was decided. It was inconclusive: the second setting's wall exceeded the default's by $+0.1719$ on average, against a forecast of $+0.1953$, and only one problem reached $+0.125$; at threshold $0.1$ the mean was $+0.1289$. As the settings also differ in step size, this mean is not a reading of $\varepsilon$; at a fixed step, raising $\varepsilon$ moved the wall by $-0.3672$ at the default step and $+0.0703$ at the larger. The next test therefore changes only $\varepsilon$ and tests the model's forecasts directly.

\textbf{A second test.}
It reads the same four problems at threshold $0.05$ with another set of five seeds ($200$, $203$, $206$, $209$ and $212$), at $T$ and the new budgets $4T$ and $16T$. Its four settings differ only in $(\tilde{\eta}, \varepsilon)$: the first test's default, and the larger step with the default $\varepsilon$, with $\varepsilon = 0.01$, or with $\varepsilon$ scaled like $\tilde{\eta}$, which keeps the cap $\tilde{\eta} / \varepsilon$ at its calibration value. Of four readings fixed in advance, the first asks whether changing only $\varepsilon$ changes the wall's response to the budget from $4T$ to $16T$. The second compares the model with every wall, also at threshold $0.1$. The third tests the law for $\varepsilon^\ast$ in twelve conditions where only $\varepsilon$ moves: $\rho$, the zero-crossing ratio, is the zero of a line fitted to the error below $\varepsilon^\ast$, in units of $\varepsilon^\ast$, and $c$ is the error at $1.1\,\varepsilon^\ast$ over that at $\varepsilon \approx 0$. The fourth asks whether scaling $\varepsilon$ with the step raises the wall at $T$. Forecasts and bands were committed before any test run, and the model has no fitted constant. We fixed the model's structure and this design after seeing the calibration problem, and added the scaled setting after the model forecast that it would win; that forecast could have lost. The first test read these problems at $T$ with other seeds.

\begin{figure}[t]
\centering
\includegraphics[width=\textwidth]{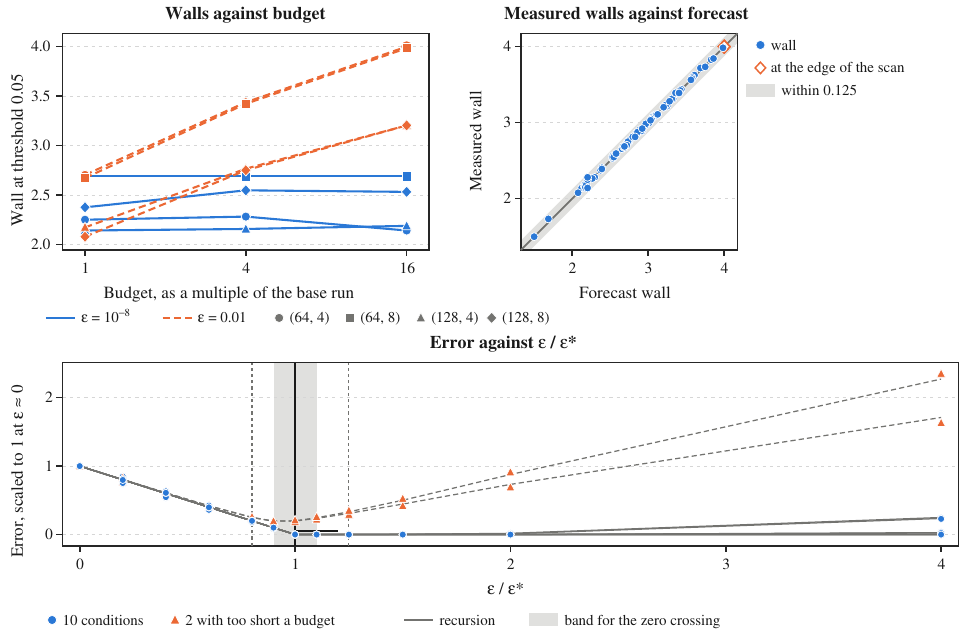}
\caption{Second zeroth-order test. Top left: walls against budget for the larger step; with $\varepsilon = 10^{-8}$ they barely move from $4T$ to $16T$, with $\varepsilon = 0.01$ they rise. Hollow marks reached the end of the scan. Top right: every measured wall against its forecast, with the tolerance band. Bottom: error after $4T$ steps, scaled to $1$ at $\varepsilon \approx 0$, with the recursion's curves, the zero-crossing band, its outer limits (dotted) and the limit on $c$ (short bar). Forecasts and bands were registered before any test run.}
\label{fig:zo}
\end{figure}

\begin{table}[ht]
\centering
\small
\caption{The four readings of the second zeroth-order test (first and last crossings coincide on all $88$ walls).}
\label{tab:zo}
\begin{tabular}{@{}p{0.42\textwidth}llp{0.27\textwidth}@{}}
\toprule
Reading & Forecast & Measured & Band \\
\midrule
Change of the wall from $4T$ to $16T$, $\varepsilon = 0.01$ minus $10^{-8}$ at the larger step & $+0.51$ & $+0.54$ & $\ge +0.25$; $\ge +0.125$ on three problems \\
Walls within $0.125$ of the forecast, thresholds $0.05$ and $0.1$ & -- & $88$ of $88$ & $\ge 80$; none beyond $0.5$ \\
Conditions with $\rho \in [0.9, 1.1]$; collapses with $c \le 0.05$ & $12$; $10$ & $12$; $10$ of $10$ & $\ge 11$; $\ge 9$ of $10$ \\
Gain at $T$ from scaling $\varepsilon$, against $\varepsilon = 0.01$; against the default setting & $+0.23$; $+0.43$ & $+0.24$; $+0.43$ & second $\ge +0.25$; both $\ge +0.125$ on three problems \\
\bottomrule
\end{tabular}
\end{table}

\textbf{Outcome.}
All four readings held (Table~\ref{tab:zo}, Figure~\ref{fig:zo}). The second reading's tolerance, $0.125$, is half a grid step, not a measured seed difference; it shows only that every wall lies within half a step of its forecast. In the third reading, ten conditions test the closed form: $\rho$ runs from $0.960$ to $1.013$ and the largest $c$ is $0.0003$. The other two ($V = 128$, $s = 3.0$) have, by the model's account, too short a budget and test only the recursion: $\rho$ is $1.034$ and $1.061$, against $1.057$ and $1.052$. A calibration check stopped this test as well: on the calibration problem one of two zero-crossing ratios read $1.1023$, just outside the band. We fixed beforehand what a repeat would decide. With $60$ new seeds the ratios read $1.0138$ and $0.9876$, five-seed readings scattered by about $0.04$, and the stopping reading sat at the high edge of that scatter. Band and statistic did not change. Scaling $\varepsilon$ alone gained under a grid step on average but over half a step on every problem; against the default the gain includes the step change. This covers one exact quadratic (exact two-point differences, sign vectors, no minibatch noise, a $\beta_2$ window longer than the strongest modes' mixing time), these four problems and, outside the second reading, threshold $0.05$; we claim nothing about $\varepsilon^\ast$ in real models. We wrote the design, model, bands and outcome; forecasts and bands predate every test run, a separate automated agent ran the committed plan, and a program decided.

\textbf{What is new.}
We have not seen the floor or $\varepsilon^\ast$ in prior work. Normalized least mean squares adds a constant $\varepsilon$ to $\varepsilon + \|x(n)\|^2$ to avoid division by zero~\citep{rathnakara2017new}, whereas variable step-size LMS algorithms~\citep{ramadan2004performance, ramadan2005robust, park2009variable, zhao2025} (the error-data step formula of \citealp{ramadan2005robust} is reproduced in \citealp{zipf2007estudo}, with a simplified variant in \citealp{rathnakara2017new}) vary a scalar gain $\mu(n)$ over time from error and input magnitudes or bounded error nonlinearities rather than setting a fixed denominator constant in Adam's second-moment recursion from the curvature spectrum. Putting a scalar preconditioner into the experimental relation of \citet{song2026zeroth} gives $\sqrt{\hat{v}} = \tilde{\eta}\,\mathrm{tr}\,H / 2 - \varepsilon$, a step that paper does not take; this is our expression without the momentum correction, which lowers $\varepsilon^\ast$ by $0.87\%$ to $4.19\%$ on the test conditions (within the third reading's band). We add the closed-recursion derivation and a test with forecasts committed before its runs.

\section{Main Result: Breaking the Wall by a Forecast Amount}
\label{sec:break}

Sections~\ref{sec:step} and~\ref{sec:zo} found that one constant sets the wall ($\eta / c$ in first order, $\varepsilon$ against $\varepsilon^\ast$ in zeroth-order Adam). Here we set it by the rule on new problems against the same-step default, with forecasts, bands and reading code committed beforehand (no fitted constant), run by a separate agent and scored once (Tables~\ref{tab:break}--\ref{tab:axes}).

\begin{table}[ht]
\centering
\caption{Seed-median walls at threshold $0.05$ for the rule's and default constants at the same step size (Adam, $T = 60000$; zeroth order, $4T$; lift is rule minus default). Under last crossing all $8$ walls are unchanged; under first crossing $1$ of $8$ moves, by $0.3303$.}
\label{tab:break}
\begin{tabular}{llcccc}
\toprule
Method & Problem & Forecast & Measured & Default & Lift \\
\midrule
Adam & $(24, 8)$ & $5.3438$ & $5.2279$ & $3.2725$ & $1.9554$ \\
Adam & $(32, 10)$ & $4.8130$ & $4.7384$ & $3.5903$ & $1.1481$ \\
Zeroth-order Adam & lattice, $(64, 8)$ & $3.7405$ & $3.7527$ & $2.6312$ & $1.1215$ \\
Zeroth-order Adam & power law, $(32, 8)$ & $1.9521$ & $1.9572$ & $1.2296$ & $0.7276$ \\
\bottomrule
\end{tabular}
\end{table}

\textbf{First order.}
At $U^\ast$ ($U^\ast {U^\ast}^\top = M^\ast$), the Gauss--Newton Hessian $H = (2/m) J^\top J$ ($J_{k,:} = \mathrm{vec}(2 A_k U^\ast)^\top$) has the $D(D-1)/2$-dimensional gauge nullspace $\{U^\ast \Omega : \Omega^\top = -\Omega\}$ from $U^\ast \mapsto U^\ast Q$, and $h_{\min}, h_{\max}$ are its extreme eigenvalues on the orthogonal complement~\citep{pw2026a}; when a small constant leaves a strong mode's step above $L = 2(1+\beta_1)/(1-\beta_1)$, that mode enters a period-2 flip oscillation whose measurement coupling floors the weak modes~\citep{pw2026b}. On matrix sensing at two new shapes, $(24, 8)$ with ten seeds and $(32, 10)$ with five ($T = 60000$), the rule sets $c = \eta\, h_{\max} / 36$ on each instance ($A = 36$) from the true solution's curvature, for Adam (at $A = 36$ and $30$) and grafted Shampoo. These walls all lie below the committed forecast, by $0.1159$, $0.1164$ and $0.0849$ on $(24, 8)$ and $0.0746$, $0.0786$ and $0.0609$ on $(32, 10)$. Bands reach further below than above, without adjusting the forecasts. Adam with the default constant and $\eta$ lowered until $A < 38$ falls on the same rule ($0.2578$ below on $(24, 8)$; $0.1460$ and $0.2322$ below on $(32, 10)$): the rule depends only on $\eta / c$.

At $\eta = 0.01$, moving the constant from $10^{-8}$ to the rule's value raises Adam's wall by $1.9554$ on $(24, 8)$ and $1.1481$ on $(32, 10)$, and grafted Shampoo's by $1.5937$ and $1.7425$, rising on every seed and at $0.02$ and $0.1$.

Once the constant dominates, Adam's update is momentum gradient descent with the same step. Run directly, taking the highest wall over three warm-ups, momentum gradient descent reaches the same wall (the rule's wall minus this one: $-0.0565$ on $(24, 8)$ and $-0.0360$ on $(32, 10)$). Without warm-up every run failed, as anticipated. Setting the step ratio $\eta / c$ in advance, not Adam's normalization, raises the wall.

The stability limit sits at $38$. On $(24, 8)$, at $A \le 36$ the error fell below $10^{-9}$ in all $60$ runs of each of Adam, momentum gradient descent and grafted Shampoo, and at $A = 50$ and $60$ it stayed above in $35$, $40$ and $28$ of $40$. Past the limit a larger $A$ lowers the wall: at $A = 100$ and $300$ the walls are $4.2556$ and $4.0421$.

The rule's setting is not the highest we tried. Adam with $\eta = 0.01$ decayed to $0$ on a cosine schedule and the default constant reaches a wall $0.5181$ higher on $(24, 8)$ and $0.2597$ higher on $(32, 10)$, on every seed. Thus the rule's constant from the true curvature raises the wall by a forecast amount over the unscheduled same-step default, while a scheduled default goes higher.

\begin{table}[ht]
\centering
\caption{Axes covered (yes) or left out (no), Sections~\ref{sec:break}--\ref{sec:beyond}.}
\label{tab:axes}
\small
\begin{tabular}{lcc}
\toprule
Axis & First order & Zeroth order \\
\midrule
Seeds, shapes or spectra, budgets, thresholds & yes & yes \\
Longer budgets & no & yes \\
Step size, default constant & yes & yes \\
Step size, rule's/bound's constant & no & yes \\
Multiple of $\varepsilon^\ast$ & -- & yes \\
Constant without true curvature/spectrum; $\mathrm{Var}(t_k)$ fix, $>4T$ or non-quadratic $\hat{t}$ & yes; -- & yes; no \\
Cosine schedule with rule's constant & yes & -- \\
Momentum decays $\beta_1$; $\beta_2$ & yes; no & no \\
Measurement ratio, preconditioner refresh & no & -- \\
Initial scale, numerical precision & no & -- \\
Hadamard sign rotation; general rotation & -- & yes; no \\
Perturbation law; difference step; minibatch noise & -- & no; yes; no \\
Other families, non-quadratic losses & yes & yes \\
Handwritten digits ($B = 32, 256, n$); other $B$/$A$/$\beta_1$, $n$, $\lambda$, networks, datasets & yes; no & no \\
Logit scale $\rho$; joint $\rho,\beta_1$, $\varepsilon \in (10^{-8}, c)$, other $d$/$A$, minibatches, digits & yes; no & no \\
Queries $q$ ($\tilde{\eta}$ fixed); unscaled $\eta$, other $q$/$d$, non-quadratic, minibatches & -- & yes; no \\
\bottomrule
\end{tabular}
\end{table}

\textbf{Zeroth order.}
We turn the Hessians of two quadratics by a Hadamard rotation with random signs: the lattice problem has the curvature of Section~\ref{sec:zo} on $(64, 8)$ and $T = 80000$, and the power-law problem has $\lambda_k = k^{-s}$, $k = 1, \dots, VD$, on $(32, 8)$ and $T = 40000$; both use five seeds and budgets $T$ and $4T$. Under this rotation the mean-square model of Section~\ref{sec:zo} gains one term, $\mathbb{E}[(q_i^\top a z)^2 \mid e] = (1 - 2/(VD)) \|He\|^2 + 2\lambda_i^2 \tilde{e}_i^2$ in the eigenbasis $\tilde{e} = Q^\top e$, where the exact fourth-moment identity holds for any orthogonal $Q$ (Appendix~\ref{app:prop}); the closed recursion in the mode energies needs $|Q_{ji}| = 1/\sqrt{VD}$, which the Hadamard sign rotation gives, shifting the denominator from $1 + \kappa\gamma\lambda_i$ to $1 - \gamma\lambda_i/(1+\beta_1)$. The rule keeps $\eta = 0.003$ and sets $\varepsilon = 1.1\,\varepsilon^\ast$, where $\varepsilon^\ast = \tilde{\eta} / \gamma^\ast$ and $\gamma^\ast$ now solves
\[
\frac{\gamma}{2} \Bigl(1 - \frac{2}{VD}\Bigr) \sum_i \frac{\lambda_i}{1 - \gamma \lambda_i / (1 + \beta_1)} = 1 .
\]

The model, committed before any run, places the walls: on each problem all $14$ walls lie within $0.125$ of it, the farthest by $0.0690$ and $0.0782$. The walls at the three multiples of $\varepsilon^\ast$ we ran, and of the fixed-step method below, all lie slightly above the model; walls below it occur only under the default $\varepsilon$. Changing only $\varepsilon$, from $10^{-8}$ to $1.1\,\varepsilon^\ast$ at the same $\eta$, raises the wall at budget $4T$ by $1.1215$ and $0.7276$; the walls at thresholds $0.02$ and $0.1$ rise too ($2.0708$ and $0.5531$; $1.3687$ and $0.4002$), and so do all five seeds.

With the rule's $\varepsilon$ the wall moves with the budget, and with the default it does not. From $T$ to $4T$ the wall with the rule's $\varepsilon$ climbs from $2.8163$ to $3.7527$ and from $1.5762$ to $1.9572$, while the default's stays at $2.6631$ and $2.6312$, and at $1.2505$ and $1.2296$; the floor is removed under the rotation as well. With the default $\varepsilon$, lowering $\eta$ as the budget grows also catches up: the default step with the highest wall is $0.003$ at $T$ and $0.001$ at $4T$ on the lattice problem, and $0.001$ and $0.0003$ on the power-law problem. The rule's $\varepsilon$ needs no retuning as the budget changes.

Once the constant dominates, the method is momentum zeroth-order descent with a fixed step, and that method, run directly, reaches the same wall (the rule's wall minus this one at $T$ and $4T$: $-0.1160$ and $+0.1768$ on the lattice problem, $-0.0017$ and $+0.0800$ on the power-law problem). Beyond the wall the normalization still acts: at three points past it, the tail error at $4T$ is $0.974$, $2.61$ and $3.46$ with the fixed step against $0.240$, $0.367$ and $0.463$ with the rule's $\varepsilon$ on the lattice problem, and $6.01$, $4.28$ and $7.00$ against $0.676$, $0.774$ and $0.835$ on the power-law problem.

The value $\varepsilon^\ast$ lies where the extended model puts it. Zero crossings read from a scan of $\varepsilon$ fall at $0.9901$ and $0.9940$ of the model's zero on the lattice problem, where the diagonal recursion of Section~\ref{sec:zo} cannot be told apart from it, and at $0.9097$ and $0.9326$ on the power-law problem, where the diagonal recursion puts the zero at $0.8168$ and $0.7473$; only the second of these readings tells the two models apart.

This covers one Hadamard sign rotation, two quadratics, five seeds, two budgets, and a constant from the spectrum.

\textbf{Cost.}
Neither lift costs accuracy on the strongest directions or in the loss. At the first, third and fifth points of each grid, the error on those directions and the loss under the rule, over those under the default, are at most $4.7 \times 10^{-5}$ and $8.4 \times 10^{-6}$ on the two matrix sensing shapes and $0.000412$ and $1.77 \times 10^{-6}$ on the two quadratics. Section~\ref{sec:further} continues these tests.

\section{The Constant from Data, the Stability Limit, and the Cosine Schedule}
\label{sec:further}

Forecasts and bands predate the runs (Table~\ref{tab:further}); walls are $0.05$ medians ($\eta = 0.01$, $T = 60000$; $(24, 8)$, $(32, 10)$).

\textbf{The constant from the measurements.}
From the measurements alone, $\hat{M} = V \sum_k b_k A_k$ has mean $M^\ast$, top $D$ eigenpairs $\hat{U}$, and top Gauss--Newton eigenvalue $\hat{h}$ at $\hat{U}$ replacing $h_{\max}$ in $c = \eta\, \hat{h} / 36$. The wall meets the rule for the resulting $A$, $5.1896$ vs.\ $5.2295$ and $4.6694$ vs.\ $4.6849$, as with $h_{\max}$ ($5.2987$ vs.\ $5.3684$, $4.7837$ vs.\ $4.8037$), rising above the default by $1.6755$, $1.1431$ ($2.6232$, $1.5329$ at $0.02$; $1.7057$, $1.1889$ at $0.1$) on every seed. Since $\hat{h} / h_{\max} \in [1.188, 1.433]$ exceeds $1$ on all instances, $c$ is conservative at the precomputed cost: $0.1092$, $0.1143$ below the $h_{\max}$ wall ($0.1387$, $0.1188$ computed). This estimator is specific to this sensing family ($10$ and $5$ seeds, $A = 36$; starting curvature, computed not run, puts $A$ past the limit on some instances).

A zeroth-order default pilot at budget $T$ that averages $\sqrt{\hat{v}}$ over the last tenth gave $0.9926$--$1.0174\,\varepsilon^\ast$ on the lattice problem (five-seed medians per $s$) but read low on the power-law one from $s = 1.625$, down to $0.7903$. From $201$ starting loss values ($K = 100$) estimating $\mathrm{tr}\,H$ as $\hat{t}$, $\hat{\varepsilon} = \tilde{\eta}\hat{t}/2$ puts the $1.1\,\hat{\varepsilon}$ and $1.1\,\varepsilon^\ast$ walls within $0.125$ at $4T$ and $0.05$ across four quadratics (Table~\ref{tab:b13}). Under rotation the step-limit correction changes sign: on $\mathrm{R2}$ the ratio $\hat{\varepsilon}/\varepsilon^\ast$ reads low ($1.1\,\hat{\varepsilon} < \varepsilon^\ast$ on all instances), the $0.02$ floor returns at $4T$, and the recursion caps the $16T$ wall below the $1.1\,\varepsilon^\ast$ wall ($16T$ model only, not run); the noiseless upper bound $2.2\,\hat{\varepsilon} = 1.1\,\tilde{\eta}\hat{t}$, below $\varepsilon^\ast$ on none of $135$ instances, costs little on $\mathrm{R2}$ at $0.05$ and more on the other three cells.

\begin{table}[t]
\centering
\caption{Starting-trace constant $\hat{\varepsilon} = \tilde{\eta}\hat{t}/2$ ($1.1\hat{\varepsilon}$ and $2.2\hat{\varepsilon}$) minus $1.1\varepsilon^\ast$ at $4T$ ($K = 100$, five seeds; $\mathrm{R1}$: Hadamard-rotated lattice quadratic $(64, 8)$; $\mathrm{R2}$: Hadamard-rotated power-law quadratic $(32, 8)$). All $36$ walls lie within $0.125$ of recursion; on $\mathrm{R2}$, $\hat{\varepsilon}/\varepsilon^\ast$ medians are $0.5672$--$0.8149$ and $16T$ $0.05$ walls are $1.9105$ vs.\ $2.2401$ (model only, not run).}
\label{tab:b13}
\footnotesize
\begin{tabular}{lcccc}
\toprule
Cell & $1.1\hat{\varepsilon} - 1.1\varepsilon^\ast$ ($0.05$) & Model & $0.02$ gap & $2.2\hat{\varepsilon} - 1.1\varepsilon^\ast$ ($0.05$, model) \\
\midrule
$(48, 4)$ & $-0.0264$ & $-0.0289$ & $-0.0279$ & $-0.3366$ ($-0.3460$) \\
$(16, 12)$ & $-0.0845$ & $-0.0557$ & $-0.0322$ & $-0.5103$ ($-0.5530$) \\
$\mathrm{R1}$ & $+0.0216$ & $+0.0115$ & $+0.0173$ & $-0.2806$ ($-0.2727$) \\
$\mathrm{R2}$ & $-0.0027$ & $-0.0554$ & $-0.1454$ ($-0.2254$) & $-0.0593$ ($-0.0621$) \\
\bottomrule
\end{tabular}
\end{table}

\textbf{The stability limit moves with momentum.}
The limit follows $2(1 + \beta_1)/(1 - \beta_1)$, the exponential-moving-average momentum threshold~\citep{cohen2022adaptive}: $18$ at $\beta_1 = 0.8$ and $78$ at $0.95$, so $38$ holds only at $\beta_1 = 0.9$. At $\beta_1 = 0.8$, all runs fell below $10^{-9}$ at $A \in \{11.37, 14.21, 17.05\}$, while $18$, $20$, $20$, $19$ of $20$ Adam runs stayed above at $A \in \{18.95, 20.84, 23.68, 28.42\}$; at $0.95$, all runs fell below at $A \in \{49.26, 61.58, 73.89\}$ (all $> 38$), while $18$, $20$, $20$, $20$ stayed above at $A \in \{82.11, 90.32, 102.63, 123.16\}$, as did every warmed-up momentum run above the limit (read on $(24, 8)$, $s \in \{2, 3\}$, $20$ runs per $A$, one budget, cutoff $10^{-9}$).

\textbf{The cosine schedule.}
On new seeds the rule's wall falls on its forecast ($0.0721$, $0.0813$ below); the cosine-scheduled default is $0.2901$ higher on $(24, 8)$ ($9/10$ seeds), while on $(32, 10)$ the rule catches up ($0.0721$ lower, higher on $3/10$).

With a small constant, Adam's normalization holds strong directions while weak ones, once $c$ dominates, run momentum descent at step $\eta(t)/c > 36 / h_{\max}$. On earlier seeds the gap opens in the last third, where their error falls as $\exp(-h_{\min} \int \eta\, dt / c)$ plus residual $r$ ($0.1573$, $0.1905$). On new seeds walls lie $0.1685$, $0.1746$ below the model-plus-$r$ forecast and $-0.0112$, $+0.0160$ from the model ($r$ did not recur); last-third decay is $0.8731$, $1.1567$ times the model's, with pre-$50000$ share $0.7292$, $0.8595$ tracking $\int \eta$ ($0.8697$), not one half.

Multiplying $c$ by $10$ lowers the wall by $0.5504$, $0.2680$ and dividing by $10$ raises it by $0.7484$, $0.7508$ on all seeds; on $(24, 8)$ the default's gain comes from $c$, not cosine, rising to $10^{-9}$ ($-0.0886$, $+0.0906$ vs.\ model).

With the rule's $c$, the cosine halves $\int \eta\,dt$ and the wall falls at the rule's value for $T/2$ ($-0.1363$, $-0.1323$), sitting $0.6936$, $0.4317$ below the cosine default and $0.4036$, $0.3596$ below the rule. At fixed $\eta$, small $c$ leaves strong directions above $L = 2(1 + \beta_1)/(1 - \beta_1)$, where oscillation floors weak directions~\citep{pw2026b}, whereas the rule's large $c$ keeps all steps below $L$; under $\eta(t) \to 0$, strong directions drop below $L$ late while small $c$ gives weak directions $\int \eta\, dt / c > 36 T / h_{\max}$, using opposite ends of $c$ ($\eta_0 = 0.01 \to 0$).

\section{Longer Budgets, a Non-Quadratic Loss, and Logistic Regression}
\label{sec:beyond}

Forecasts and bands predate the runs (Table~\ref{tab:further}); walls are seed medians at $0.05$ ($\beta_1 = 0.9$; $q = 1$ unless noted).

\begin{table}[t]
\centering
\caption{Walls at $0.05$ across Sections~\ref{sec:further}--\ref{sec:beyond}: forecast, rule, default, and lift (rule minus default). Of $43$ walls ($38$ printed), all $38$ printed are unchanged under last crossing; under first crossing $8$ move, up to $1.8350$ ($6$ start above $0.05$, where first upward and last crossings coincide).}
\label{tab:further}
\vspace{2pt}
\small
\setlength{\tabcolsep}{3.5pt}
\begin{tabular}{@{}>{\raggedright\arraybackslash}p{0.24\textwidth}>{\raggedright\arraybackslash}p{0.10\textwidth}ccc>{\centering\arraybackslash}p{0.16\textwidth}@{}}
\toprule
Study & Problem & Forecast & Rule & Default & Lift \\
\midrule
Matrix sensing, $\hat{h}$, Adam ($\eta = 0.01$, $T = 60000$) & $(24, 8)$ / $(32, 10)$ & $5.2295$ / $4.6849$ & $5.1896$ / $4.6694$ & $3.5140$ / $3.5263$ & $+1.6755$ / $+1.1431$ \\
Matrix sensing, cosine ($\eta_0 = 0.01 \to 0$) & $(24, 8)$ / $(32, 10)$ & $5.3428$ / $4.8334$ & $5.2707$ / $4.7521$ & $5.5607$ / $4.8242$ & $-0.2901$ / $-0.0721$ \\
\addlinespace
Diagonal quadratic, $64T$, ZO Adam ($\eta = 0.003$) & $(48, 4)$ / $(16, 12)$ & $5.4263$ / $8.5177$ & $5.4247$ / $8.5004$ & $2.3388$ / $3.5701$ & $+3.0859$ / $+4.9303$ \\
Diagonal quadratic, $16T$, vs.\ $\eta = 0.001$ ($q = 4$; $16$) & $(48, 4)$ / $(16, 12)$ & $4.7012$ / $7.3999$ & $4.7100$ / $7.4567$ & $4.3221$ / $6.8820$ & $+0.3879$ / $+0.5747$; $-0.0578$ / $-0.0872$ \\
\addlinespace
Matrix sensing, $16T$, ZO Adam ($\eta = 0.003$, $\mu = 10^{-3}$) & $(24, 8)$ & $2.9835$ & $2.9227$ & $< 0$ & $\ge +2.9227$ \\
Matrix sensing, $16T$, ZO Adam ($\eta = 0.003$, $\mu = 10^{-4}$ vs.\ $10^{-3}$) & $(24, 8)$ & $3.5178$ & $3.5850$ & -- & $+0.6623$ \\
\addlinespace
Population logistic, Adam ($\eta = 0.01$, $T = 60000$) & $d = 64$ / $128$ & $3.5013$ / $2.9700$ & $3.5042$ / $2.9746$ & $3.5901$ / $3.0622$ & $-0.0859$ / $-0.0876$ \\
Population logistic, $16T$, ZO Adam ($\eta = 0.003$, $\mu = 10^{-3}$) & $d = 64$ / $128$ & $2.8098$ / $2.5162$ & $2.8118$ / $2.5151$ & $1.8350$ / $1.6377$ & $+0.9768$ / $+0.8774$ \\
Digits logistic, full-batch, Adam ($\eta = 0.01$) & $r = 16$ / $32$ & $3.9864$ / $3.2544$ & $4.0086$ / $3.3587$ & $4.3017$ / $3.5854$ & $-0.2930$ / $-0.2266$ \\
Digits logistic, $B = 32$ / $256$, Adam ($r = 16$) & $B = 32$ / $256$ & $1.8136$ / $2.3298$ & $1.8023$ / $2.3052$ & $1.8029$ / $2.3076$ & $-0.0006$ / $-0.0024$ \\
Digits logistic, $B = 32$ / $256$, Adam ($r = 32$) & $B = 32$ / $256$ & $1.6748$ / $2.1021$ & $1.6611$ / $2.0932$ & $1.6619$ / $2.0962$ & $-0.0008$ / $-0.0029$ \\
\bottomrule
\end{tabular}
\end{table}

\begin{figure}[t]
\centering
\includegraphics[width=\textwidth]{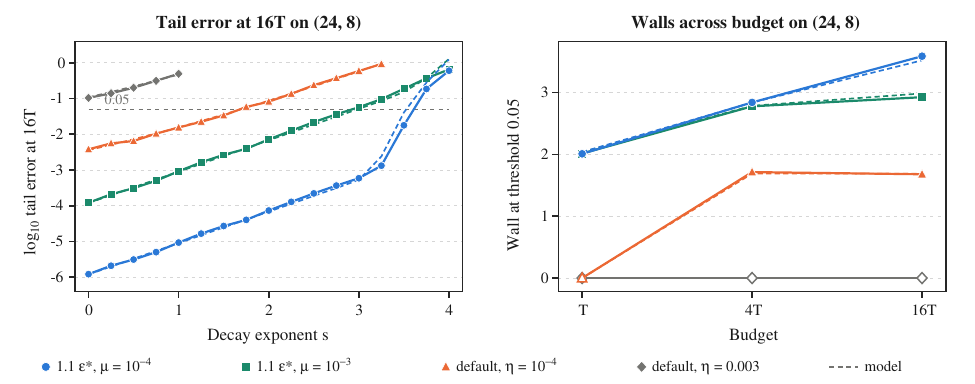}
\caption{Two-stage floor in zeroth-order Adam on matrix sensing at $(24, 8)$. Left: median tail error at $16T$ against $s$ with model curves (dashed) and $0.05$ threshold; default $\varepsilon = 10^{-8}$ ($\eta = 0.003$) sits on the noise floor above $0.05$, $\varepsilon = 1.1\,\varepsilon^\ast$ ($\mu = 10^{-3}$) meets the cubic floor at $2.9227$, and $\mu = 10^{-4}$ moves the $16T$ wall to $3.5850$. Right: measured (solid) and model (dashed) walls across $T$, $4T$, $16T$ (hollow: below grid).}
\label{fig:twostage}
\end{figure}

\textbf{Budgets up to $64T$.}
On two new Section~\ref{sec:zo} shapes, $(48, 4)$ and $(16, 12)$ (five new seeds, $\eta = 0.003$, budgets $T$ to $64T$), the model placed all $16$ walls of each within $0.125$. Under default $\varepsilon$ the wall stops at the floor: at $64T$ it is $2.3388$ and $3.5701$ against $2.2228$ and $3.6346$, where the floor of \eqref{eq:floor} meets the threshold, and $3.5822$ and $5.7983$ against $3.5444$ and $5.7739$ at $\eta = 0.001$; from $16T$ to $64T$ these move by $-0.0943$ to $+0.0392$. On $(48, 4)$ the $\eta = 0.003$ default exceeds the floor at every budget ($+0.1160$ at $64T$). Smaller $\eta$ lowers the floor and raises the cap. Under the rule's $\varepsilon$ the wall rises $2.3490$ to $4.9312$ from $T$ to $64T$, and at $64T$ sits where an error decaying at the step $\gamma^\ast / 1.1$ reaches the threshold: $5.4247$, $8.5004$ against $5.4263$, $8.5177$. Changing $\varepsilon$ adds $2.2907$ and $3.8523$ to that rise, and at $64T$ the rule leads by $3.0859$ and $4.9303$ on all five seeds. At $q = 1$, the best default has $\eta = 0.001$; the rule exceeds it at $T$, $4T$, $16T$, $64T$ by $0.6600$, $0.8184$; $0.4499$, $0.6776$; $1.1981$, $1.7759$; $1.8425$, $2.7021$, using the spectrum for $\varepsilon^\ast$. For $q \in \{4, 16\}$ queries per step on these $d = 192$ shapes (five new seeds; $16T$ queries, $KT/q$ steps), $\tilde{\eta}$ keeps its $q = 1$ value ($\varepsilon^\ast(q)$ uses the spectrum). At $16T$ the model placed all six $(48, 4)$ walls within $0.125$ ($+0.0040$--$+0.0585$) and five of six on $(16, 12)$ ($-0.1101$--$+0.1431$, none $> 0.25$); from $q = 4$ to $16$, the $\eta = 0.003$ default rises $0.8209$, $1.5946$ (model $0.8676$, $1.3882$), $\varepsilon = 1.1\,\varepsilon^\ast(q)$ moves $-0.2197$, $-0.4032$ (model $-0.2259$, $-0.3671$; steps alone: $-0.8841$, $-1.3743$), and against the $\eta = 0.001$ default the rule leads by $+0.3879$, $+0.5747$ at $q = 4$ while at $q = 16$ that default catches up ($-0.0578$, $-0.0872$; model $-0.0575$, $-0.1033$).

\textbf{A non-quadratic loss in zeroth order.}
On matrix sensing at $(24, 8)$ (five Section~\ref{sec:break} instances, $\mu = 10^{-3}$, budgets to $16T$; Figure~\ref{fig:twostage}), zeroth-order Adam at $\eta = 0.003$ and default $\varepsilon$ sits on its noise floor at $16T$: at $s = 0$, $0.25$, $0.5$ its error is $0.9638$, $0.9172$, $0.9173$ times the model's floor, and at $T$, $4T$, $16T$ at least $0.1$ across the grid. At $\eta = 10^{-4}$ the floor drops and the $16T$ wall is $1.6773$ (model $1.6876$). The rule's $\varepsilon$ removes the noise floor and stops on a second floor, $2.9227$ at $16T$ (model $2.9835$): the quartic loss in $U$ has a two-point cubic term from the third derivative at the truth that leaves a floor proportional to $\mu^2$ (Appendix~\ref{app:prop}). At $\mu = 10^{-4}$ the $16T$ wall rises $0.6623$ at $0.05$ and $1.0523$ at $0.02$ (model $0.5343$, $0.8896$), while $T$ and $4T$ nearly coincide at $0.05$. Changing $\varepsilon$ raises the $16T$ wall by $\ge 2.9227$ on all five seeds (default $> 0.05$), exceeding the best default ($\eta = 10^{-4}$ at $4T$, $16T$) by $1.0623$, $1.2454$, and $\ge 2.0123$ at $T$. The local model at the truth placed all $12$ walls within $0.25$ ($4$ below the grid in both, $8$ within $0.0672$). This covers one shape and two $\mu$ with $\varepsilon^\ast$ from the truth's Hessian; MeZO in Appendix~\ref{app:failed} is not explained here.

\textbf{Logistic regression, first order.}
With Gaussian design $x \sim \mathcal{N}(0, S)$ ($S$ eigenvalues $k^{-s}$, Haar eigenvectors), true-model logits $x^\top w^\ast$ of variance $9$, and population loss minimized at $w^\ast$, we start at $w = 0$ for $d = 64$ (ten seeds) and $128$ (five) and read relative error on $S$'s weakest quarter. At $\eta = 0.01$ and $T = 60000$, $c = \eta\, h_{\max} / 36$ ($A = 36$) gives $3.5042$ and $2.9746$ against the default's $3.5901$ and $3.0622$ ($-0.0859$, $-0.0876$; negative at $0.02$, $0.1$ and on every seed). Without the truth, $c = \eta\, \lambda_{\max}(S/4) / 36$ from the Hessian bound $S/4$~\citep{bohning1988monotonicity} gives $3.0771$ and $2.6351$ ($-0.5130$, $-0.4271$ vs.\ default). Across four default $\eta$ and a cosine schedule, $\eta = 0.01$ is highest at both $d$, within $0.25$ of the rule. On ten new seeds at $d = 64$ ($\beta_2 = 0.999$, read at $20000$, $40000$, $60000$) and four shapes---logit SD $\rho \in \{1, 6\}$ at $\beta_1 = 0.9$, $\beta_1 \in \{0.8, 0.95\}$ at $\rho = 3$---the $\eta = 0.01$ default wall at $T$ sits where $a\, h_{\max}$ meets $L = 2(1 + \beta_1)/(1 - \beta_1)$, forecast in advance ($a = 0.999\, L / h_{\max}$): $3.4130$, $3.6991$, $3.4114$, $3.7771$ against $3.3953$, $3.7116$, $3.3933$, $3.7712$, within $0.0181$ ($L$~\citep{cohen2022adaptive} is empirical for Adam). It rises $0.3656$ from $\beta_1 = 0.8$ to $0.95$ ($0.3779$ from $L$) and $0.2862$ from $\rho = 1$ to $6$ ($0.3163$ forecast); $\eta = 0.003$ minus $0.01$ default is $-0.0174$ to $-0.0238$. With $c = \eta\, h_{\max} / A$ ($A = 36 L / 38$, truth's Hessian), the rule meets the short-form band ($-0.0121$--$+0.0217$) and lies $-0.0751$ to $-0.0956$ below the $\eta = 0.01$ default at $0.05$: the default already sits at $L$, so $A$'s $36$-to-$38$ margin cannot raise it (the gap exceeds the full recursion's $-0.0130$--$-0.0146$; digits not covered).

\textbf{Logistic regression, zeroth order.}
With $\mu = 10^{-3}$, five seeds, budgets to $16T$ and $\varepsilon^\ast$ from the truth's Hessian, changing only $\varepsilon$ at $\eta = 0.003$ raises the $16T$ wall by $0.9768$ and $0.8774$ ($0.6256$, $0.5110$ at $0.1$; $\ge 2.7318$, $\ge 2.4475$ at $0.02$, default below grid) on all five seeds. The rule's wall rises to $1.4235$, $2.3254$, $2.8118$ at $T$, $4T$, $16T$ and $1.4464$, $2.1206$, $2.5151$, while the default stalls by $4T$ ($1.5194$, $1.8242$, $1.8350$; $1.5230$, $1.7300$, $1.6377$), so changing $\varepsilon$ adds $1.0726$ and $0.9540$ from $T$ to $16T$. The best of three default step sizes lies within $0.25$ of the rule at each budget ($-0.0958$, $+0.0520$, $+0.0323$; $-0.0766$, $+0.0673$, $+0.0760$). The local model at the truth from $w = 0$ placed $10$ of $12$ walls within $0.25$ at $d = 64$ ($4T$, $16T$ defaults $+0.2726$, $+0.2833$) and all $12$ at $d = 128$ (within $-0.1638$). Where the model puts the error on its floor by $16T$, the $16T$ defaults sit on it at $d = 128$ (medians $1.0019$, $0.9852$, $1.0286$ across the three $\eta$) and at $0.8421$, $0.8741$, $0.8827$ at $d = 64$ (points $0.6494$--$1.1481$). Zero crossings from a scan of $\varepsilon$ ($s = 1.5$, $2$) fall at $0.9279$, $0.8497$ and $0.9075$, $0.8943$ of $\varepsilon^\ast$, against $1.0000$, $1.0000$, $1.0000$, $1.0035$ from the recursion.

\textbf{Multi-class logistic regression on handwritten digits, first order.}
Ten seeds draw $1000$ of $1797$ digits ($64$ pixels, $10$ classes; $797$ test), center, whiten and project onto $r = 16$ or $32$ principal modes ($144$ or $288$ parameters) with $x_s = M x_w$, $M = Q\,\mathrm{diag}(q)$, $q_j = (j+1)^{-s/2}$, Haar $Q$, and penalty $(\lambda/2)\|M^\top\Theta\|^2$ ($\lambda = 0.001$); Newton finds the truth to $\le 10^{-13}$, and Adam from $\Theta = 0$ ($\beta_2 = 0.999$) runs to $T = 60000$ (read at $20000$, $40000$, $60000$ by last crossing of $0.05$; first and last crossings coincide in all $144$ full-batch readings). Full-batch at $\eta = 0.01$, $c = \eta\, h_{\max} / 36$ from the truth puts the wall on forecast, $4.0086$ and $3.3587$ vs.\ $3.9864$ and $3.2544$ ($+0.0223$, $+0.1044$), below the $\eta = 0.01$ default ($-0.2930$, $-0.2266$ at $0.05$; negative at $0.02$, $0.1$ and on every seed). Theorem~2.1(b) of \citet{bohning1992multinomial} (Theorem~5.3 of \citet{bohning1988monotonicity}) bounds the Hessian via the design second moment, giving $\hat{h} = \tfrac{1}{2} + \lambda$ and $c = \eta\, \hat{h} / 36$ without the truth: at $r = 16$ the wall meets forecast, $2.9579$ vs.\ $2.8299$ ($+0.1280$), while $r = 32$ is undecided ($2.4247$ vs.\ $2.2506$, $+0.1741$); near the wall $\kappa = \hat{h} / h_{\max}$ is $18.9$--$32.2$ and $27.8$--$37.7$, with cost $1.0507$, $0.9340$ vs.\ $1.1564$, $1.0038$ forecast. Across five defaults ($0.1$--$0.001$) and cosine, $\eta = 0.1$ leads ($4.4361$, $3.6563$), $2.1599$ and $1.8467$ below a rough travel cap ($6.5961$, $5.5030$); rule and bound sit $0.4275$, $0.2975$ and $1.4782$, $1.2315$ lower (negative at $0.02$, $0.1$ and on all seeds). At $T$, on $s = 2.5, 3, 3.5$ and $1.75, 2.25, 2.75$, head-error and loss-gap ratios to the $\eta = 0.01$ default are $\le 1$ under the rule, and under the bound reach $17.54, 191$ and $2497, 1.943 \times 10^5$ on $s = 3, 3.5$ ($\le 1$ at $2.5, 1.75$), while test loss stays $0.2106$--$0.2111$ and $0.1601$--$0.1602$. Arithmetic bounds $\eta = 10^{-5}$ ($A < 38$ at default $c$) to $\le 0.6$ per coordinate in $T$ steps, below the $s = 0$ truth ($1.919$--$3.297$). With without-replacement batches $B \in \{32, 256\}$ on ten new seeds ($130$--$139$, $(r, B) \in \{16, 32\} \times \{32, 256\}$, last crossing of $0.05$ at $T = 60000$; Table~\ref{tab:minibatch}), gradient-noise covariance is $C = S\,(n - B)/[B(n - 1)]$, and the linearized momentum recursion with batch-curvature feedback (Appendix~\ref{app:minibatch}, spectral radius $1$ at $A^\ast(B)$) sets $\sqrt{\hat{v}}$ within $1.25\times$ at $\eta = 0.01$, $0.1$ and places all $12$ walls of the $\eta = 0.01$ default, $\eta = 0.01$ rule $c = \eta\, h_{\max} / 36$, and $\eta = 0.1$ default within $0.125$ of forecast ($-0.0774$ to $+0.0002$): gradient noise sets the default step and wall at $\eta = 0.01$, $0.1$, while the $\eta = 1$ default stops at the left grid edge ($1.4198$--$1.5297$ below forecast). Because $c = \eta\, h_{\max} / 36 \ll \sqrt{\hat{v}}$ at $\eta = 0.01$, the rule minus default is $-0.0029$ to $-0.0006$ (per-seed $-0.0032$ to $-0.0003$), removing the full-batch gap. Scaling $\eta$ and $c$ to $c = 1$ above $\sqrt{\hat{v}}$ with a $1000$-step linear warm-up and $\eta = A_M / h_{\max}$ ($A_M = 2, 11$ at $B = 32, 256$) also stops at the left grid edge in all four cells, below forecast and below the $\eta = 0.01$ default, while $\eta = 0.1$ leads $\eta = 0.01$ by $0.6258$--$0.8469$ and $c = 1$ by $0.9045$--$1.2877$ (forecast within $0.25$ of top default). Across the three checkpoints ($\ge 9$ seeds dropping in both intervals, only the first, or neither; Table~\ref{tab:minibatch}), every $c = 1$ curve has a left plateau and right falling branch, with $60000$-step minima $0.1133$--$0.1488$ ($6$--$29\times$ the unfeedback floor, all $> 0.05$); across five batch streams per seed at those minima ($s = 3, 2.25, 2.75, 2$; cross-seed means do not share $\Theta^\ast$, so we split within each seed), median bias $e_{\mathrm{bias}} = 0.1428, 0.0876, 0.1486, 0.1328$ exceeds stream SD $e_{\mathrm{var}} = 0.0331, 0.0680, 0.0233, 0.0869$ by $e_{\mathrm{bias}} / e_{\mathrm{var}} = 4.3708, 1.3320, 6.5508, 1.4796$ ($10, 9, 10, 10$ seeds; last-$1000$-step time SD/mean $0.0207, 0.2371, 0.0067, 0.3007$). Nearest the $c = 1$ forecast wall, $A = 36$ at $c = 1$ gives head errors $> 10\times$ $A_M$, and on the last three grid points shared with the $\eta = 0.01$ default, $c = 1$ head-error and loss-gap ratios to default are $1.4269$--$2.8497$ ($1$--$10$; $> 10$ forecast). Sample-matched budgets, other $B$, $\eta$, $A$, warm-ups, $\beta_1,\beta_2$, without-truth constants, epoch shuffling, zeroth order, and networks were not read.

\begin{table}[t]
\centering
\caption{Minibatch logistic regression on digits across $(r, B)$ (seeds $130$--$139$, $T = 60000$; Appendix~\ref{app:minibatch}).}
\label{tab:minibatch}
\vspace{2pt}
\footnotesize
\setlength{\tabcolsep}{2pt}
\begin{tabular}{@{}>{\raggedright\arraybackslash}p{0.23\textwidth}*4{>{\centering\arraybackslash}p{0.182\textwidth}}@{}}
\toprule
Quantity & $(16, 32)$ & $(16, 256)$ & $(32, 32)$ & $(32, 256)$ \\
\midrule
$\sqrt{\hat{v}}$ / model ($\eta = 0.01$ / $0.1$) & $0.9987$ / $0.9631$ & $1.0006$ / $0.9873$ & $1.0008$ / $0.9663$ & $1.0009$ / $0.9817$ \\
$\eta = 0.01$ default (forecast) & $1.8029$ ($1.8142$) & $2.3076$ ($2.3322$) & $1.6619$ ($1.6756$) & $2.0962$ ($2.1049$) \\
$\eta = 0.01$ rule (forecast) & $1.8023$ ($1.8136$) & $2.3052$ ($2.3298$) & $1.6611$ ($1.6748$) & $2.0932$ ($2.1021$) \\
Rule $-$ $\eta = 0.01$ def.\ (fc.) & $-0.0006$ ($-0.0006$) & $-0.0024$ ($-0.0024$) & $-0.0008$ ($-0.0008$) & $-0.0029$ ($-0.0028$) \\
$\eta = 0.1$ default (forecast) & $2.5743$ ($2.6517$) & $3.1545$ ($3.1557$) & $2.2877$ ($2.3613$) & $2.7534$ ($2.7533$) \\
$\eta = 1$ default (forecast) & $1.7500$ ($3.1946$) & $2.2500$ ($3.6698$) & $1.2500$ ($2.7797$) & $1.7500$ ($3.1713$) \\
$A^\ast(B)$ median ($A_M(B)$) & $3.708$ ($2$) & $20.847$ ($11$) & $4.009$ ($2$) & $22.049$ ($11$) \\
$c = 1, A_M$ (forecast; gap) & $1.5000$ ($3.0750$; $-1.5750$) & $2.2500$ ($3.7363$; $-1.4863$) & $1.0000$ ($2.6196$; $-1.6196$) & $1.7500$ ($3.1504$; $-1.4004$) \\
$c = 1$ $-$ $\eta = 0.01$ def.\ (fc.) & $-0.3029$ ($+1.2608$) & $-0.0576$ ($+1.4041$) & $-0.6619$ ($+0.9440$) & $-0.3462$ ($+1.0455$) \\
Head $A=36/A_M$ ($s$); $c=1$/def. & $29.3992$ ($3$); $2.8497$ & $11.2442$ ($3.75$); $2.2054$ & $44.0203$ ($2.5$); $1.8037$ & $30.7608$ ($3.25$); $1.4269$ \\
$c=1$ flat ($s$; $60\text{k}/20\text{k}$) & $6$ ($1.50$--$2.75$; $0.8635$--$1.0012$) & $5$ ($2.25$--$3.25$; $0.9770$--$1.3297$) & $6$ ($1.00$--$2.25$; $0.9289$--$1.0568$) & $6$ ($1.75$--$3.00$; $0.6946$--$1.3185$) \\
$c=1$ settle; fall ($s$) & $3.00$; $8$ ($\ge 3.25$) & $3.50$; $9$ ($\ge 3.75$) & $2.50$; $8$ ($\ge 2.75$) & $3.25$; $8$ ($\ge 3.50$) \\
$c=1$ $60\text{k}$ min ($s$) / floor (ratio) & $0.1459$ ($3.00$) / $0.0233$ ($6.2571$) & $0.1177$ ($2.25$) / $0.0064$ ($18.3143$) & $0.1488$ ($2.75$) / $0.0184$ ($8.0935$) & $0.1133$ ($2.00$) / $0.0039$ ($28.8621$) \\
\bottomrule
\end{tabular}
\end{table}

\section{Discussion}
\label{sec:discussion}

\textbf{What sets the ungrafted wall.}
Once damping dominates, Section~\ref{sec:step} reads the ungrafted wall as relaxation at rate $(\eta / \sqrt{\varepsilon})\, h_{\min}$. Over $1035$ earlier rows decaying between steps $20000$ and $60000$, the median measured-to-model rate is $1.03$--$1.07$ on the four $D \le 6$ shapes and $1.14, 1.23$ on $D = 8$ (two weakest modes). Extrapolating to $t = 0$ gives median $\ln e_0$ of $0.154$--$0.321$ on $D \le 6$ ($0.1541$ on $(12, 4)$, $0.2140$ on $(16, 6)$, mean $0.1841$) and $0.329, 0.504$ on $D = 8$, whereas \eqref{eq:step} sets $\ln e_0 = 0$, matching the detection lag in \citet{pw2026a}. Replacing $\ln 20$ by $\ln 20 + 0.1841$ (registered forecasts fixed) cuts the $(14, 5)$ $32$-wall rms gap from $0.1561$ to $0.1191$ ($0.1637$ to $0.1266$ Shampoo, $0.1481$ to $0.1112$ Adam; max $0.2155$ to $0.1780$), removing $0.2648$ of mean gap $-0.1494$ ($32$ below): per-run $20000$--$60000$ secant $\ln e_0$ cuts mean and rms gaps to $-0.0493$ and $0.0838$ ($22$ below, $0.6703$ removed: $+0.0095$, rms $0.0449$ on the $14$ walls at $T \le 60000$; $-0.0949$, rms $0.1045$ at $T \ge 120000$, where $84/180$ bracketing runs had $e_{20000} \ge 1$). Adding per-run rates leaves mean $-0.0069$, rms $0.0088$ at $T \le 60000$ and overshoots at $T \ge 120000$ ($32$-wall mean $+0.0646$, rms $0.1298$); $-\ln\rho(F)$ moves walls by $\le 0.0011$ (share $-0.0025$; $\sqrt{\hat{v}}$ unlogged).

Adam at $10^{-8}$ runs above $A = 38$, losing this relaxation; \citet{cohen2022adaptive} give Eq.~(1) for RMSProp and Adam without testing Shampoo or SOAP, while our post-hoc readings meet $38$ in Shampoo and match it in SOAP, and $\varepsilon_g = \sqrt{\varepsilon}$ restores the ungrafted median wall ($0.10$ on the mean from one seed).

\textbf{What the main result shows.}
In both orders fixed-step momentum at the rule's step reaches the rule's wall, so the constant acts via the step, not normalization. Two step thresholds---$L = 2(1 + \beta_1)/(1 - \beta_1)$ in first order and $\varepsilon^\ast = \tilde{\eta}/\gamma^\ast$ in zeroth order---separate the four regimes in Table~\ref{tab:regimes}; where the rule raises the wall and where it does not follow from one model, which also forecasts the default walls at the logistic stability limit and under minibatch noise in Table~\ref{tab:minibatch} (within $0.25$ of the highest default).

\begin{table}[ht]
\centering
\caption{Four regimes of the denominator constant across Sections~\ref{sec:main}--\ref{sec:beyond}.}
\label{tab:regimes}
\vspace{2pt}
\footnotesize
\setlength{\tabcolsep}{3.5pt}
\begin{tabular}{@{}>{\raggedright\arraybackslash}p{0.22\textwidth}>{\raggedright\arraybackslash}p{0.24\textwidth}c>{\raggedright\arraybackslash}p{0.39\textwidth}@{}}
\toprule
Regime & Condition & Sections & Consequence for $c$ or $\varepsilon$ \\
\midrule
Fixed-step oscillation floor & Small default $c$ leaves strong modes above $L$ & \ref{sec:main}, \ref{sec:step}, \ref{sec:break}, \ref{sec:further} & $c = \eta h_{\max}/36$ (or $\eta\hat{h}/36$) keeps modes below $L$, raising the wall as forecast. \\
\addlinespace
Cosine time-integrated step & $\eta(t) \to 0$ drops modes below $L$ late; small $c$ enlarges $\int\eta\,dt/c$ & \ref{sec:break}, \ref{sec:further} & Cosine default beats rule on $(24, 8)$, matches on $(32, 10)$; rule's $c$ halves $\int\eta\,dt/c$, lowering the wall. \\
\addlinespace
Full-batch stability-limit saturation & Default sits at $a h_{\max} \approx L$ or larger $\eta$ is stable & \ref{sec:beyond} & The $36$-to-$38$ margin cannot lift same-step default on logistic regression; $\eta = 0.1$ is higher on digits. \\
\addlinespace
Minibatch and zeroth-order floors & Noise sets $\sqrt{\hat{v}} \gg c$; zeroth-order variance ($< \varepsilon^\ast$) and cubic ($\propto \mu^2$) floors & \ref{sec:zo}, \ref{sec:break}, \ref{sec:beyond} & Noise buries $c = \eta h_{\max}/36$ under $\sqrt{\hat{v}}$ and stalls $c = 1$; in zeroth order $1.1\,\varepsilon^\ast$ removes the variance floor (leaving sensing's $\mu^2$ floor), while retuned $\eta$ catches up on rotated quadratics, population logistic, and $q = 16$. \\
\bottomrule
\end{tabular}
\end{table}

\textbf{Practical advice, within these limits.}
\begin{enumerate}
    \item \emph{Report both matrix damping and grafting denominator for Kronecker preconditioners.} A shared damping can reverse rankings; changing only $\varepsilon_g$ moved the median wall by $1.26$ and $1.83$ on average.
    \item \emph{Vary step size and sample count separately in zeroth-order scaling.} Tying them hid the Evolution Strategies gain.
    \item \emph{Report median and mean together across seeds.} They differed in the Shampoo study.
\end{enumerate}

\section{Related Work}
\label{sec:related}

\textbf{Kronecker preconditioners and grafting.}
Kronecker-factored methods approximate curvature or gradient covariance with tensor factors, as in K-FAC~\citep{martens2015optimizing}, E-KFAC~\citep{george2018fast} and Shampoo~\citep{gupta2018shampoo,anil2020scalable,shi2023distributed}. \citet{anil2020scalable} found that ill-conditioned factor matrices lose precision when inverse roots are computed, which led them to double-precision roots. Instability from widely spread per-layer step sizes led them to layer-wise grafting~\citep{agarwal2020disentangling} onto a first-order optimizer. \citet{shi2023distributed} separate the matrix eigenvalue shift from the grafting denominator in their distributed implementation. \citet{vyas2025soap} showed that, in idealized form, power-$1/2$ Shampoo equals Adafactor~\citep{shazeer2018adafactor} in the preconditioner's eigenbasis and introduced SOAP, and \citet{morwani2025new} analyzed the same form as one step of Kronecker power iteration while setting the damping aside. Closest to our Shampoo study, \citet{eschenhagen2025purifying} decoupled Shampoo's eigenvalue and eigenbasis updates. On image classification, ungrafted Shampoo underperformed AdamW and often diverged when its roots were updated every $100$ steps; updated every step, it matched grafted Shampoo's final loss on a vision transformer but not on ConvNeXt V2, and per-step eigenvalue correction removed the need for grafting. They noted, without testing it, that the best learning rate may transfer across optimizers when the effective damping is matched. \citet{lin2025understanding} found that power-$1/2$ Shampoo fails on RWKV7 language models without grafting, unlike their KL-Shampoo. Muon~\citep{jordan2024muon,liu2025muon} takes a different route, orthogonalizing updates on hidden layers while keeping embeddings and heads on AdamW.

\textbf{Damping through a root.}
\citet{eschenhagen2026clarifying} read the damping through a root before we did: they compare the best damping of two Shampoo exponents through each exponent's own root of $\varepsilon$, and state that the damping in the matrix root is distinct from the constant in Adam grafting. We measure a different conversion. With both factors at exponent $-1/4$ their roots multiply to $\sqrt{\varepsilon}$, so we set the grafting constant to that value, changing nothing else, and read the wall against the ungrafted arm on two controlled problems, comparing only these two values. Two citation indexes list $180$ works citing the four papers nearest this comparison~\citep{anil2020scalable,shi2023distributed,eschenhagen2025purifying,eschenhagen2026clarifying}. We scanned the $132$ with full text on arXiv: $29$ mention grafting and $7$ compare runs with and without it, none with the grafting constant at a root of Shampoo's damping. Both indexes are incomplete, so it may still have been done.

\textbf{Weak directions in low-rank recovery.}
That strong directions are learned first is classical. In deep linear networks with orthogonal inputs, a mode's learning time is inversely proportional to its strength~\citep{saxe2014exact}. From a small start, gradient flow on matrix factorization learns the target's components one by one in order of eigenvalue~\citep{li2021towards}, and gradient descent on matrix sensing learns solutions of increasing rank~\citep{jin2023understanding}; \citet[Theorem~3.3]{stoger2021small} bound its iterations on a target $XX^\top$ by a term inversely proportional to the step size and the square of $X$'s smallest singular value, up to a logarithm. We take this order as known and measure its consequence at a fixed budget. Scaled gradient descent from a spectral initialization converges at a rate independent of the condition number~\citep{tong2021accelerating}. From a small random start the damped form of \citet{xu2023power} acts like gradient descent, and near the solution the damping's effect looks negligible, the reverse of our runs, where it takes over late. \citet{zhang2021preconditioned} view the damping as interpolating between scaled and plain gradient descent, and \citet{zhang2024fast} let it decay with the error, noting that the speed-up lasts only until the error reaches the damping's order; ours stays fixed.

\textbf{Damping as a step size.}
\citet{sabanayagam2026optimizer} derive mode-by-mode laws for deep matrix factorization as continuous-time flows under full observation, with balanced and aligned factors. There a mode of instantaneous Adam whose gradient is far below $\varepsilon$ behaves, as an approximation, like gradient descent up to a factor $1/\varepsilon$, and cumulative Shampoo has the denominator $\sqrt{\lambda + h(t)}$, with $\lambda$ the accumulators' initial value and $h$ the gradient history, so every mode starts in a phase like gradient descent and a mode near its target settles exponentially at a rate that falls as this denominator grows. Our reading of the ungrafted wall has the same form, with three differences: our $L$ and $R$ forget and $\varepsilon$ is added at every inversion, so the damping takes over late in a run and stays; we use matrix sensing with unaligned factors and read each instance's own curvature; and we run at finite step size with momentum, where a stability limit appears. \citet{pw2026a} use the smallest non-gauge eigenvalue at leading order, and \citet{pw2026b} read a wall as a finite-budget crossover for Adam at small constants and use the limit $38$; both work on a family with power-law measurement covariance.

\textbf{Hyperparameters and ranking reversals.}
\citet{choi2019empirical} inverted or collapsed optimizer rankings by tuning more hyperparameters, including the denominator constant of adaptive methods; they view that constant as interpolating between diagonal preconditioning and momentum SGD. \citet{schmidt2021descending} treat Adam tuned in its learning rate alone and Adam tuned in all hyperparameters as different optimizers. AlgoPerf~\citep{dahl2023benchmarking,kasimbeg2025accelerating} links an algorithm's performance to its search space and tuning budget, and language-model benchmarks~\citep{semenov2025benchmarking,wen2025fantastic} found that hyperparameter choices change the ranking of optimizers such as AdamW, SOAP and Muon. \citet{eschenhagen2026clarifying} report that tuning the damping shifts the ranking of the optimizers they compare. \citet{singh2026loss} sort nine update rules on underdetermined matrix sensing at a budget of $2 \times 10^4$ steps and call the gap they find a statement about that budget and size range; one rule closes it when run longer.

\textbf{Power-law spectra and zeroth-order optimization.}
Scaling under power-law eigenvalue decay has been analyzed for first-order methods~\citep{bordelon2024dynamical,paquette2024four}, and \citet{kunstner2024zipf} found that under heavy-tailed class imbalance gradient descent is slow on low-frequency classes while Adam and sign descent are less sensitive. For gradient-free optimization, \citet{nesterov2017random} established how Gaussian random gradient estimators scale with dimension, and \citet{malladi2023fine} showed that two-point MeZO converges at a rate governed by the local effective rank of the Hessian when fine-tuning pretrained models. Evolution Strategies~\citep{salimans2017evolution,lange2022evosax} use the same kind of estimate at population scale. ZO-AdaMM~\citep{chen2019zo} and HiZOO~\citep{zhao2024second} combine such estimators with diagonal moment or Hessian preconditioning, and sparse perturbations or coordinate-wise estimators~\citep{liu2018zeroth,zhang2024revisiting} reduce the variance. Our work adds a direct measurement of where a rare direction is no longer recovered within the budget, using the same kind of threshold crossing for first-order, second-order and zeroth-order methods.

\section{Limitations}
\label{sec:limitations}

\begin{enumerate}
    \item \textbf{Scale.} All studies are full-batch except the handwritten-digits run at $B \in \{32, 256\}$, with $V \le 128$ and $D \le 12$, $d \in \{64, 128\}$ on population logistic regression, or $r \in \{16, 32\}$ on handwritten digits, with five or ten seeds. Nothing here should be carried to minibatch language-model training, low-precision arithmetic or deep networks without measuring there.
    \item \textbf{Raising the wall.} The first-order constant came from the curvature at the truth, from an estimate ready-made only for this sensing family, or from a bound on the curvature; on logistic regression neither constant raised the wall, and on handwritten digits a larger default step size was higher.
    \item \textbf{Estimating $\varepsilon^\ast$.} A pilot average missed the power-law spectrum; the starting-trace constant holds at $4T$ on four quadratics, but its $\mathrm{R2}$ ratio reads low and its $16T$ wall is capped lower (model-only).
    \item \textbf{Logistic regression.} We read only a Gaussian population loss (four follow-up shapes at $d = 64$, two $\eta$, bands at $T = 60000$) and, on handwritten digits, two shapes, one dataset and one $\lambda$ with coordinate ill-conditioning, full-batch and at $B \in \{32, 256\}$ drawn without replacement (fixed step budget $T = 60000$, truth's $h_{\max}$, model's $A_M$); sample-matched budgets, other $B$ or $A$, epoch shuffling and zeroth order on the digits were not read.
    \item \textbf{One cosine schedule.} It was read in one form, decayed to $0$ from one initial step size.
    \item \textbf{Two test problems.} The Shampoo study's grafting results rest on two problems, five seeds and one step size, budget, refresh interval and root.
    \item \textbf{One reference.} The comparison with Adam rests on one configuration. Five of the eleven run in the first test would put its reading in the failed range, so we do not rank grafted Shampoo against Adam.
    \item \textbf{Median against mean.} Both grafting tests read the median; on the mean neither would meet its rule.
    \item \textbf{No shift reported.} The crossing shift changes with seeds, problem and aggregate.
    \item \textbf{The denominator test.} It reads walls and not runs, compares two denominator values, and its average leans on the three lowest levels.
    \item \textbf{Process.} We wrote the qualifications in Section~\ref{sec:main} after each reading, several checks behind them were unplanned, and the first test had no valid independent review before it ran. We set the second test's rule and read its outcome ourselves; the rule was committed first, and a program computed the outcome.
    \item \textbf{Coarse grids.} The truncation and Evolution Strategies walls use a grid with step $0.5$, so a half-step error could make those outcomes inconclusive, though not confirmed.
    \item \textbf{Untested factors.} We tested the ratio of measurements to degrees of freedom on one shape and budget, and not cutting only the left matrix, given that $U$ is unique only up to a right rotation. The population formula is an underived heuristic, and MeZO's failure to converge (Appendix~\ref{app:failed}) may reflect only its shorter effective horizon.
    \item \textbf{The zeroth-order tests.} The first two use one exact quadratic with sign vectors, four problems and threshold $0.05$; the query test read two $d = 192$ shapes, $q \in \{4, 16\}$, five seeds ($q = 1$ on different seeds). The proposition needs the same setting; its step statement is mean-field. We claim nothing about $\varepsilon^\ast$ in real models, where minibatch noise needs a new derivation.
    \item \textbf{An early look.} In the first zeroth-order test, after the forecasts, bands and decision code were fixed but before the runs finished, the model's author, testing a check, ran six medians with the test seeds on $(64, 4)$; no file entering the decision changed afterwards.\label{sec:endmain}
\end{enumerate}

{\small
\setlength{\bibsep}{1.5pt}
\label{sec:endbib}
}

\appendix

\section{The Three Earlier Studies}
\label{app:failed}

\begin{table}[h]
\centering
\caption{Walls from the three failed studies on the problem $(V, D) = (16, 4)$, $m = 96$, read from the mean over five seeds.}
\label{tab:failed}
\vspace{2pt}
\small
\begin{tabular}{>{\raggedright\arraybackslash}p{0.2\textwidth}>{\raggedright\arraybackslash}p{0.5\textwidth}>{\raggedright\arraybackslash}p{0.21\textwidth}}
\toprule
Study & Walls & Change tested \\
\midrule
Shampoo to SOAP & Shampoo $4.3744$, Eigen-Adafactor $3.5899$, SOAP $3.6140$; Adam on the unrotated problem $3.6178$ & $-0.7605$ along the path \\
\addlinespace
Truncating SOAP's left matrix & Block size 16: $3.7859$; 8: $3.9947$; 4: $4.0826$; 2: $4.0327$; 1: $3.9649$ & $-0.1789$ from size 16 to size 1 \\
\addlinespace
Evolution Strategies & 4 pairs: $1.5184$; 128 pairs under the schedule: $1.4486$; 128 pairs at the 4-pair step size: $1.8216$ & Measured over predicted gain: $-0.1676$; at fixed step size, $+0.7275$ \\
\bottomrule
\end{tabular}
\end{table}

The first study bisected its crossings, and the other two interpolated on a grid with step $0.5$. In the third study the power law sits in the measurements and not in the target: $M^\ast = U^\ast {U^\ast}^\top$, where $U^\ast$ has independent $\mathcal{N}(0, 1/V)$ entries, $A_k = \Sig^{1/2} (G_k + G_k^\top) \Sig^{1/2} / 2$ with $G_k$ as in Section~\ref{sec:setup} and $\Sig \propto \mathrm{diag}(1^{-s}, \dots, V^{-s})$ of trace $V$, and the wall is read from the relative error of the whole matrix, $\|U_T U_T^\top - M^\ast\|_F / \|M^\ast\|_F$. Its prediction came from a heuristic formula in which the wall grows with half the logarithm of the number of queries per step, divided by $\ln(VD)$. The fixed-step run was an unscored check. In the same study two-point MeZO did not converge even at $s = 0$, so its wall is $0$.

\section{The First Grafting Test}
\label{app:first}

\begin{figure}[h]
\centering
\includegraphics[width=0.92\textwidth]{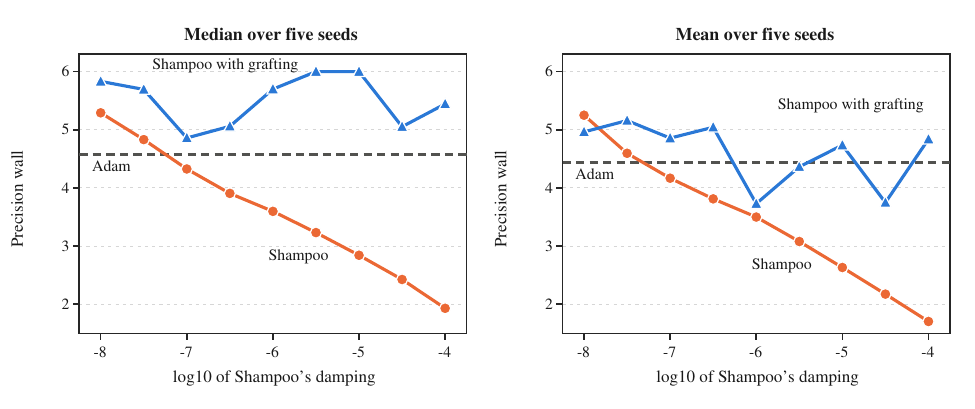}
\caption{First test: precision wall of Shampoo against its damping on $(V, D) = (12, 4)$. Left: median over five seeds. Right: mean over the same seeds. Blue triangles: Shampoo with grafting. Orange circles: Shampoo. Dashed line: Adam at step size $0.01$ and constant $10^{-8}$. A wall of $6$ (scan edge) means no crossing was found.}
\label{fig:grafting}
\end{figure}

Figure~\ref{fig:grafting} shows the first test on $(12, 4)$, with bisection narrowing each wall to a bracket of width $0.0234375$. On its five seeds the grafted median wall stays above the reference at all nine levels, by at least $0.2838$, and the mean wall first falls below it at $-6.0$. Neither statement carried over to the second test (Section~\ref{sec:main}), and the two tests' grids differ, so their walls should not be compared point by point.

\section{Derivation of Proposition~\ref{prop:step}}
\label{app:prop}

Let $m$ be Adam's first moment before bias correction, so that one step is $e' = e - \gamma\,c\,m'$ with $c = 1 / (1 - \beta^t)$ and $\beta = \beta_1$, where primes mark the next step. Let $B_i = \mathbb{E}[e_i m_i]$ and $C_i = \mathbb{E}[m_i^2]$. On a pure quadratic the two-point difference does not depend on the perturbation size, and $a = \sum_i \lambda_i e_i z_i$. The sign vector is independent of the past, so $a z_i$ has conditional mean $\lambda_i e_i$. Since $z_i^2 = 1$ its square is $a^2$, and the cross terms drop out of $\mathbb{E}[a^2] = G^2$. With $X_i = \mathbb{E}[e_i m_i']$,
\begin{align*}
C_i' &= \beta^2 C_i + 2 \beta (1 - \beta) \lambda_i B_i + (1 - \beta)^2 G^2, \\
X_i &= \beta B_i + (1 - \beta) \lambda_i A_i, \\
A_i' &= A_i - 2 \gamma c X_i + \gamma^2 c^2 C_i', \\
B_i' &= X_i - \gamma c\, C_i'.
\end{align*}
These lines are exact for any fixed sequence of steps, and they close because $G^2$ depends only on the $A_i$.

At a constant step, once $c = 1$, a stationary state with $G > 0$ has $B_i = -\gamma C_i / 2$ and $C_i = (1 - \beta) G^2 / (1 + \beta + \beta \gamma \lambda_i)$, which gives \eqref{eq:floor}. Substituting \eqref{eq:floor} into $G^2 = \sum_i \lambda_i^2 A_i$ and cancelling $G^2$ gives \eqref{eq:gstar}.

For any $\beta \in [0, 1)$ the threshold is the same. At a constant step with $c = 1$, writing $m_t = m'_t$, one mode obeys $(e_{i,t+1}, m_{i,t})^\top = F_i\,(e_{i,t}, m_{i,t-1})^\top + v_i\,\xi_{i,t}$ with $\xi_{i,t} = a_t z_{i,t} - \lambda_i e_{i,t}$, $F_i = \bigl(\begin{smallmatrix} 1 - \gamma(1-\beta)\lambda_i & -\gamma\beta \\ (1-\beta)\lambda_i & \beta \end{smallmatrix}\bigr)$ and $v_i = \bigl(-\gamma(1-\beta),\, 1-\beta\bigr)^\top$. The spectral radius of $F_i$ is below $1$ on $0 < \gamma \lambda_i < 2(1+\beta)/(1-\beta)$. Since $\xi_{i,t}$ has conditional mean $0$ given the history and conditional second moment $\sum_{j \ne i} \lambda_j^2 e_{j,t}^2$ in the diagonal model, or $(1 - 2/d)\sum_j \lambda_j^2 \tilde{e}_{j,t}^2 + \lambda_i^2 \tilde{e}_{i,t}^2$ under Section~\ref{sec:break}'s Hadamard rotation (written unifiedly as $\sum_j K_{ij} \lambda_j^2 A_{j,t}$ with $K_{ij} \ge 0$), unrolling the linear system from $t = 0$ and taking expectations gives the renewal equation $A_i(t) = \mathrm{free}_i(t) + \sum_{s=0}^{t-1} h_i(t - 1 - s) \sum_j K_{ij} \lambda_j^2 A_j(s)$, where $\mathrm{free}_i(t) \ge 0$ is the unforced response and $h_i(\ell) = ((F_i^\ell v_i)_1)^2 \ge 0$ is the squared impulse response. The mass $S_i := \sum_{\ell \ge 0} h_i(\ell)$ is the stationary $A_i$ of one mode under unit-variance noise: there $G^2 = 1 + \lambda_i^2 A_i$ in the four lines, so the stationary relation behind \eqref{eq:floor}, at step $\gamma$, gives $A_i = \alpha_i (1 + \lambda_i^2 A_i)$ and $S_i = \alpha_i / (1 - \alpha_i \lambda_i^2)$ with $\alpha_i = \gamma / [2\lambda_i(1 + \kappa \gamma \lambda_i)]$. Because the kernel $h_i(\ell) K_{ij} \lambda_j^2$ is nonnegative, the renewal system contracts to zero if and only if the Perron root of the total-mass matrix $M_{ij} = S_i K_{ij} \lambda_j^2$ is below $1$, and diverges when it exceeds $1$. An eigenvector $u > 0$ of $M u = u$ satisfies $(1 + \sigma S_i \lambda_i^2) u_i = k\, S_i \sum_j \lambda_j^2 u_j$ with $(\sigma, k) = (1, 1)$ (diagonal) or $(-1, 1 - 2/d)$ (Hadamard); dividing by $1 + \sigma S_i \lambda_i^2$, multiplying by $\lambda_i^2$ and summing over $i$ reduces $\rho(M) = 1$ to \eqref{eq:gstar} in the diagonal case and to Section~\ref{sec:break}'s equation in the Hadamard case. Every $S_i$ rises with $\gamma$, so $\rho(M)$ does too, and so do those left-hand sides, under the rotation up to the pole at $\gamma \lambda_{\max} = 1 + \beta$; hence $\rho(M) < 1$ exactly when $\gamma < \gamma^\ast$.

For $\beta = 0$ the recursion reduces to $A_i' = (1 - 2 \gamma \lambda_i) A_i + \gamma^2 G^2$. Multiplying by $\lambda_i$ and summing gives, for twice the expected loss $S = \sum_i \lambda_i A_i$, $S' = S - \gamma\,(2 - \gamma\,\mathrm{tr}\,H)\, G^2$. Since $G^2 \ge \lambda_{\min} S$, the expected loss contracts to zero when $\gamma < 2 / \mathrm{tr}\,H$ and diverges when $\gamma > 2 / \mathrm{tr}\,H$.

The same limit follows from \citet{defossez2015averaged}. With $y = H^{1/2} e$ and $X = H^{1/2} z$, a step at $\beta = 0$ is $y' = (I - \gamma X X^\top)\, y$, which is noiseless least mean squares with $\mathbb{E}[X X^\top] = H$. They define $\gamma_{\max}$ (our largest stable step) as the supremum of $\gamma > 0$ with $2\,\mathrm{Tr}(A H A) - \gamma\,\mathbb{E}[(X^\top A X)^2] > 0$ for every nonzero symmetric $A$, and note $\gamma_{\max} \le 2 / \mathrm{Tr}\,H$ for any $X$. Equality for constant norm follows: if $\|X\|^2 = \tau$ always, $(X^\top A X)^2 \le \tau\, X^\top A^2 X$, so $\mathbb{E}[(X^\top A X)^2] \le \tau\, \mathrm{Tr}(A^2 H)$ and the bound holds for $\gamma < 2 / \tau = 2 / \mathrm{Tr}\,H$, which sign vectors achieve ($\|X\|^2 = \sum_i \lambda_i z_i^2 = \mathrm{tr}\,H$).

For any orthogonal $Q$ ($H = Q\,\mathrm{diag}(\lambda)\,Q^\top$, $d = VD$, $\tilde{e} = Q^\top e$, $y = Q^\top z$, $z \in \{\pm 1\}^d$), the scalar collapse $a^2 z_j^2 = a^2$ holds exactly, and $\mathbb{E}[z_r z_s z_t z_u] = \delta_{rs}\delta_{tu} + \delta_{rt}\delta_{su} + \delta_{ru}\delta_{st} - 2\delta_{rs}\delta_{rt}\delta_{ru}$ gives $\mathbb{E}[y_i^2 y_j y_k] = \delta_{jk} + 2\delta_{ij}\delta_{ik} - 2\sum_r Q_{ri}^2 Q_{rj} Q_{rk}$ and $\mathbb{E}[(q_i^\top a z)^2 \mid e] = \|He\|^2 + 2\lambda_i^2 \tilde{e}_i^2 - 2\sum_r Q_{ri}^2 (He)_r^2$, with $(He)_r$ in parameter coordinates; for general $Q$ the last term couples $\sum_{j \ne k} \lambda_j \lambda_k \tilde{e}_j \tilde{e}_k$, so the recursion needs the full second-moment matrix rather than closing in $A_i$. Under Section~\ref{sec:break}'s Hadamard rotation ($|Q_{ri}| = 1/\sqrt{d}$), $-2\sum_r Q_{ri}^2 (He)_r^2 = -(2/d)\|He\|^2$, giving $\mathbb{E}[(q_i^\top a z)^2 \mid e] = (1 - 2/d) \|He\|^2 + 2\lambda_i^2 \tilde{e}_i^2$. Expectation over $e$ gives $(1 - 2/d) G^2 + 2\lambda_i^2 A_i$; substituting this for $G^2$ in the $C_i'$ line adds $2(1-\beta)^2 \lambda_i^2 A_i = (1-\beta)(1+\beta)\gamma\lambda_i C_i$ at stationarity ($c = 1$, $B_i = -\gamma C_i/2$), turning $(1-\beta)(1 + \beta + \beta\gamma\lambda_i) C_i$ into $(1-\beta)(1 + \beta - \gamma\lambda_i) C_i = (1-\beta)^2 (1 - 2/d) G^2$, whence $A_i = \gamma (1 - 2/d) G^2 / [2\lambda_i (1 - \gamma\lambda_i/(1+\beta))]$ and Section~\ref{sec:break}'s equation. Gaussian directions on a diagonal quadratic add $2 \lambda_i^2 e_i^2$ without $-2/d$ and share their second moment only approximately. On Section~\ref{sec:beyond}'s quartic loss $\mathcal{L}(U)$ with measurement matrices $A_k$ of Section~\ref{sec:setup}, $a(z) = \langle \nabla\mathcal{L}(U), z \rangle + r_3(z)$ has cubic term $r_3(z) = (2\mu^2/m)\sum_{k=1}^m \langle A_k, U z^\top + z U^\top \rangle \langle A_k, z z^\top \rangle$ with mean $\mathbb{E}[r_3(z)\,z] = K_\mu U$, $K_\mu = (4\mu^2/m)\sum_{k=1}^m [2 A_k^2 + D\,\mathrm{tr}(A_k) A_k - 2\,\mathrm{Diag}(A_k) A_k]$; with $H$ the Gauss--Newton Hessian at $U^\ast$ and $H^+$ its pseudoinverse on the live modes, this shifts the local fixed point at $U^\ast$ by $\Delta U^\ast = -H^+ K_\mu U^\ast \propto \mu^2$ with residual variance $\propto \mu^4$. Additive loss noise adds a term to $G^2$.

\section{Linearized Minibatch Momentum Recursion and Curvature Feedback}
\label{app:minibatch}

In Section~\ref{sec:beyond}'s minibatch handwritten-digits study ($n = 1000$, $K = 10$, $B \in \{32, 256\}$), let $\theta^\ast \in \R^{r(K-1)}$ be the Newton solution and $x = \theta - \theta^\ast$ the error in the rescaled coordinates ($x_0 = -\theta^\ast$ from $\Theta = 0$), with $M = Q\,\mathrm{diag}(q)$, $q_j = (j+1)^{-s/2}$, and $J = M \otimes I_{K-1}$. The forecast model linearizes in five steps:
(1) At $\theta^\ast$, a batch of $B$ rows drawn without replacement has gradient $g_t = (H + \delta H_t) x_t + \xi_t$, where $H = J H_w J^\top$ is the full-loss Hessian at $\theta^\ast$ (including the $\lambda$ penalty), $\xi_t$ is the batch gradient at $\theta^\ast$ with covariance $C = f S$, $f = (n - B)/[B(n - 1)]$, $S = J S_w J^\top$, $S_w = (1/n)\sum_{i=1}^n (g_i - \bar{g})(g_i - \bar{g})^\top$ (cross-entropy row gradients at $\theta^\ast$), and $\delta H_t$ is the zero-mean batch-Hessian fluctuation drawn independently each step.
(2) Freezing $\hat{v}_t$ at $v_k$ per coordinate (omitting moment correction) gives momentum descent with step $a_k = \eta / (\sqrt{v_k} + c)$, $D = \mathrm{diag}(a)$, and symmetrized Hessian $\tilde{H} = D^{1/2} H D^{1/2} = U\,\mathrm{diag}(w_m)\,U^\top$.
(3) Each mode evolves from $m_{-1} = 0$ as $(x_{t+1}, m_t)^\top = F\,(x_t, m_{t-1})^\top + (1 - \beta_1)(-1, 1)^\top \zeta_t$ with $F = \bigl(\begin{smallmatrix} 1 - (1 - \beta_1) w & -\beta_1 \\ (1 - \beta_1) w & \beta_1 \end{smallmatrix}\bigr)$, whose spectral radius reaches $1$ at $w = L = 2(1 + \beta_1)/(1 - \beta_1) = 38$. On the modes $H u_m = h_m u_m$ at $s_{\mathrm{ref}}$ ($3.0$ for $r = 16$, $2.5$ for $r = 32$) with scalar step $a = A / h_{\max}$ ($h_{\max} = \lambda_{\max}(H)$), the $2 \times 2$ Lyapunov equation gives the stationary variance ratio $\phi_m = a / [2 h_m (1 - a h_m / L)]$.
(4) The drive variance on mode $m$ is $\sigma_m^2 + f\sum_{m'} \nu_{mm'}^2 \mathrm{Var}_{m'}$, where $\sigma_m^2 = f\, u_m^\top S u_m$ and $\nu_{mm'}^2 = \mathrm{Var}_i(u_m^\top H_i u_{m'})$ (with $H_i$ the $i$-th row cross-entropy Hessian at $\theta^\ast$, dropping cross-correlations between modes of $x_t$).
(5) Hence $\mathrm{Var}_m = \phi_m(\sigma_m^2 + f\sum_{m'} \nu_{mm'}^2 \mathrm{Var}_{m'})$, or $(I - \mathrm{diag}(\phi)\,f\nu^2)\,\mathrm{Var} = \phi \odot \sigma^2$, which has a stationary second moment while $\rho(A) = \max|\mathrm{eig}(\mathrm{diag}(\phi)\,f\nu^2)| < 1$; $\rho(A^\ast(B)) = 1$ defines $A^\ast(B)$ per seed. With $\gamma(A) = 1 + (A/2)/(1 - A/L)$, the scalar feedback factor is $g(A) = v_{\mathrm{fb}}(A)/v_0(A)$, where $v_{\mathrm{fb}}(A) = \mathrm{mean}_m[(\sigma^2 + f\nu^2\mathrm{Var})_m \gamma(a h_m)]$ and $v_0(A) = \mathrm{mean}_m[\sigma_m^2 \gamma(a h_m)]$, interpolated in $\ln A$ on $A < 0.98\,A^\ast(B)$. For the $\eta \in \{0.01, 0.1\}$ defaults and $\eta = 0.01$ rule ($c = \eta h_{\max}/36$), $(a_k, v_k)$ solves $v_k = g(\bar{a} h_{\max})\bigl[C_{kk} + (1/a_k)\sum_m U_{km}^2 w_m\,\mathrm{sg}_m / (2(1 - w_m/L))\bigr]$ with $\mathrm{sg}_m = (U^\top D^{1/2} C D^{1/2} U)_{mm}$ (so feedback enters the coordinate fixed point only via the scalar $g(\bar{a} h_{\max})$ at $s_{\mathrm{ref}}$). For the $\eta = 1$ default (solving $(A/h_{\max})(\sqrt{v_{\mathrm{fb}}(A)} + c) = \eta$ below $A^\ast(B)$ at $s_{\mathrm{ref}}$) and the $c = 1$ settings ($A \in \{A_M(B), 36\}$, assuming $\sqrt{\hat{v}} \ll 1$ and omitting the $1000$-step warm-up), $a_k = A/h_{\max}$ is constant across $k$. On the whitened tail block (last $r/4$ rows, reached through $D^{1/2} U$ and $M^\top \otimes I_{K-1}$), the deterministic error $x_T = (F^T)_{00} x_0$ and the mode variances $[\mathrm{sg}_m / (2 w_m (1 - w_m/L))](1 - e^{-2 w_m T})$ (using $\mathrm{sg}_m$ without $f\nu^2\mathrm{Var}$ and capping $w_m$ at $0.999\,L$), both relative to the truth's tail norm, give per seed $e = (\mathrm{det}^2 + \mathrm{floor}^2)^{1/2}$; the median over seeds at each $s$ is read at the last crossing of $0.05$ in $\ln(-\ln e)$. Restoring $f\nu^2\mathrm{Var}$ in the tail-block floor gives $0.0267, 0.0068, 0.0226, 0.0043$ at the forecast wall (plateau minima $0.1459, 0.1177, 0.1488, 0.1133$ over it: $5.4652, 17.1891, 6.5777, 26.2546$) and $0.0267, 0.0481, 0.0148, 0.0348$ at the plateau's $s$ (ratios $5.4652, 2.4493, 10.0756, 3.2569$).

\end{document}